# Probabilistic Graphical Model using Graph Neural Networks for Bayesian Inversion of Discrete Structural Component States

Teng Li[a,b], Stephen Wu[c,d], Yong Huang[a,b,*], James L. Beck[e], Hui Li[a,b]

[a] Key Lab of Smart Prevention and Mitigation of Civil Engineering Disasters of the Ministry of Industry and Information Technology, Harbin Institute of Technology, Harbin 150090, China

[b] Key Lab of Structures Dynamic Behavior and Control of the Ministry of Education, School of Civil Engineering, Harbin, Institute of Technology, Harbin 150090, China

[c] The Institute of Statistical Mathematics, Research Organization of Information and Systems, 10-3 Midori-cho, Tachikawa, Tokyo 190-8562, Japan

[d] The Graduate University for Advanced Studies, SOKENDAI, 10-3 Midori-cho, Tachikawa, Tokyo 190-8562, Japan

[e] Division of Engineering and Applied Science, California Institute of Technology, CA, USA

## Abstract

The health condition of components in civil infrastructures can be described by various discrete states according to their performance degradation. Inferring these states from measurable responses is typically an ill-posed inverse problem. Although Bayesian methods are well-suited to tackle such problems, computing the posterior probability density function (PDF) presents challenges. The likelihood function cannot be analytically formulated due to the unclear relationship between discrete states and structural responses, and the high-dimensional state parameters resulting from numerous components severely complicates the computation of the marginal likelihood function. To address these challenges, this study proposes a novel Bayesian inversion paradigm for discrete variables based on Probabilistic Graphical Models (PGMs). The Markov networks are employed as modeling tools, with model parameters learned from data and structural topology prior. It has been proved that inferring this PGM produces the same probabilistic estimation as the posterior PDF derived from Bayesian inference, which effectively solves the above challenges. The inference is accomplished by Graph Neural Networks (GNNs), and a graph property-based GNN training strategy is developed to enable accurate inference across varying graph scales, thereby significantly reducing the computational overhead in high-dimensional problems. Both synthetic and experimental data are used to validate the proposed framework.

*Corresponding author.
*E-mail address:* huangyong@hit.edu.cn (Y. Huang). *Tel*: 86-15326669082.

## 1.Introduction

Civil infrastructures are complex systems composed of numerous interdependent structural components, whose operational states significantly influence the overall system performance. Factors such as material degradation, cumulative damage, and potential defects can adversely affect structural components, leading to progressive deterioration throughout their service life. As the service performance of the structure decreases, its health condition can be described by different discrete states (e.g., intact and damaged, or more categories). This not only simplifies the representation of health conditions but also provides crucial support for engineering decision-making. Inferring such discrete states for critical components through measurable system responses is crucial for developing targeted maintenance strategies and ensuring structural safety and reliability [1,2].

Over the last few decades, many methodologies have been developed to assess the health conditions of structures and their components [3-5], such as approaches based on structural dynamic characteristics (e.g., natural frequencies and mode shapes) [6,7], methods based on finite element model updating [8] and methods based on neural networks [9,10]. Some of these technologies evaluate the structural state by comparing dynamic or physical parameters with established health baselines, while others identify the severity of structural damages by extracting features from monitoring data and determine the state accordingly. However, many of current researches are confined to deterministic frameworks, which fail to adequately account for critical influencing factors such as environmental variability, model uncertainties, and data incompleteness. Essentially, inversion problems in engineering applications are ill-posed and ill-conditioned [11], as limited and noisy observational data makes it difficult to deterministically identify model parameters. In this context, Bayesian method offers a rigorous framework for addressing such issues by incorporating prior knowledge with monitoring data to constrain the solution space, while effectively quantifying uncertainties in the estimated parameters [12-14]. The posterior probabilities obtained from Bayesian inference can also guide for decision-making in engineering applications [15]. Furthermore, Bayesian methods naturally models the correlation between model parameters through their probabilistic framework, which can fully account for the dependencies between structural components.

In recent years, some methods based on Bayesian framework have been proposed for evaluating the structural health conditions [11,12,16-19]. For example, Jia et al. extracted detailed information about structural creep, stiffness, and pre-stress based on Bayesian inference framework to evaluate structural anomalies [17]; Huang et al. employed a hierarchical sparse Bayesian learning method to calculate the posterior probability density function (PDF) of stiffness parameters and estimate the probability of structural damage [13]. This type of method involves inferring the posterior PDF of relevant model parameters, and the formulation of likelihood function as well as the computation of the marginal likelihood is necessary during the inference process. In general, when the goal is to infer structural physical parameters (e.g. stiffness), the likelihood function between structural monitoring response and parameters can be expressed explicitly because of the deterministic relationship defined by the structural dynamics equation. However, when the objective shifts to inferring the states of structural components, the variables to be inferred will change from continuous model parameters to discrete states. This shift significantly complicates the construction of the likelihood function, due to the absence of well-defined mechanical principles that directly relate monitoring responses to component states [20]. In addition, civil structures generally comprise numerous interdependent components, which creates a high-dimensional parameter space that requires expensive computation to obtain the marginal likelihood [12]. Meanwhile, modeling the dependencies between structural components can be more challenging, as the commonly used covariance has a weaker ability to capture complex correlation patterns between discrete state variables compared to continuous model parameters. This phenomenon stems from a mismatch in its mathematical essence. Covariance is primarily used to measure linear and smooth covariation trend, while complex relationships involving discrete states are commonly intricate and nonlinear [21,22]. These difficulties limit the application of Bayesian methods in inferring the discrete component states.

To deal with these issues, this study proposes a novel paradigm for Bayesian inversion using PGMs [23]. The strong expressive capability of this framework makes it suitable for addressing the aforementioned challenges. By defining local interactions through potential functions, the dependencies between component states and the relationship between component states and system responses can be modeled. The structural topology prior of the structure and the monitoring data are then used to learn this model. Furthermore, the posterior PDF can be estimated through PGM inference, which eliminates the requirement of an analytical likelihood function and avoids the computation of marginal likelihood. Some advantages of PGMs in engineering applications have been shown in existing researches, the Bayesian network is used for the risk-based decision and condition assessment applications [24,25], and the undirected graphical model is employed to implement structural damage detection [26]. These studies demonstrate the powerful ability of PGMs, indicating that there is still room for exploration of in their applications.

However, the implementation of PGM framework also presents some challenges. Traditional inference algorithms such as variable elimination (VE), belief propagation (BP) [27] and MCMC [21] are subject to various limitations. The increasing number of model parameters will result in a significant decrease in accuracy and efficiency for these algorithms. Although the information propagation mechanisms [28,29] have been leveraged to develop neural network-based inference methods such as Graph Neural Networks (GNNs) and Message-Passing Neural Networks (MPNNs) [30,31], one critical challenge in practical applications is that training GNN for inferring high-dimensional models is computationally expensive, as both data acquisition and network optimization require substantial resources proportional to model size [32]. This limitation has motivated the development of size generalization capabilities in GNN architectures – a critical innovation enabling effective knowledge transfer from simple models to large-scale models while maintaining inference accuracy. Recent research in GNN generalization has predominantly focused on in-distribution (ID) generalization scenarios [33,34], where training and testing data share identical statistical distributions. However, this idealized assumption often fails to hold in real-world applications due to distributional shifts. While emerging studies have started to tackle more complex challenges such as out-of-distribution (OOD) generalization [35,36] and size generalization [37,38], the field still lacks a unified theoretical framework capable of rigorously characterizing the generalization boundaries of GNNs. Thus, further investigation in this area remains necessary.

In this study, we present a novel paradigm of Bayesian inversion for discrete-state assessment of structural components based on PGMs, and the marginal posterior PDF of Bayesian inference is obtained by PGM inference via a GNN-based algorithm. This framework uses Markov network to model the dependencies among discrete component states, with parameters learned through monitoring data and structural topology prior. Making inference on such model is proved to produce the same probabilistic estimation of component state as the posterior PDF derived from Bayesian inference. This improvement effectively avoids the challenges of establishing an analytical likelihood function and computing a high-dimensional summation for marginal likelihood. Furthermore, this study investigates how graph properties affect the size generalization performance of GNNs and develops a training strategy that enables GNNs to achieve accurate inference on high-dimensional problems with low computational overhead. For complex structural scenarios encountered in engineering applications, this approach ensures sufficient computational accuracy and efficiency, making it highly significant. Finally, both the synthetic data and experimental data are employed to validate the performance of the proposed framework.

The organization of this paper is as follows. In **Section 2**, we demonstrate the Bayesian inversion of discrete component states and its challenges, then we illustrate how to use PGMs to deal with such problem. The model learning method is also described. **Section 3** introduces the theoretical foundations of GNNs and how to use GNNs to implement graphical model inference. **Section 4** presents some synthetic experiments for investigating GNNs' size generalization capabilities. This section further provides comparative analyses of computational efficiency and inference accuracy between GNN-based and conventional algorithms. Finally, **Section 5** validates the framework on a truss structure, with the results demonstrating the effectiveness of the proposed method.

## 2. Bayesian inversion of discrete component states based on PGMs

### 2.1 Bayesian inversion of discrete component states and its challenges

Here we use the vector $\boldsymbol{\theta} = (\theta_1, \theta_2, \dots, \theta_n)^T$ to represent the state parameter composed of multiple structural component states, each $\theta_i$ in $\boldsymbol{\theta}$ is defined as a binary variable representing two possible states (intact and damaged). The goal of Bayesian inversion for discrete component states is to estimate the relative plausibility of each state within two discrete states based on measurable structural responses $\boldsymbol{D}$. According to Bayes' Theorem, the posterior PDF of the component states can be obtained by:

$$p(\boldsymbol{\theta}|\boldsymbol{D}) = \frac{p(\boldsymbol{\theta}, \boldsymbol{D})}{p(\boldsymbol{D})} = \frac{p(\boldsymbol{D}|\boldsymbol{\theta})p(\boldsymbol{\theta})}{\sum_{\boldsymbol{\theta}} p(\boldsymbol{D}|\boldsymbol{\theta})p(\boldsymbol{\theta})} \propto p(\boldsymbol{D}|\boldsymbol{\theta})p(\boldsymbol{\theta}) \tag{1}$$

where $p(\boldsymbol{D}|\boldsymbol{\theta})$ is the likelihood function, which expresses the probability of getting data $\boldsymbol{D}$ conditioned on the component state $\boldsymbol{\theta}$; $p(\boldsymbol{\theta})$ is the prior distribution, which quantifies the initial plausibility of each state defined by the value of $\boldsymbol{\theta}$; and $p(\boldsymbol{D}) = \sum_{\boldsymbol{\theta}} p(\boldsymbol{D}|\boldsymbol{\theta})p(\boldsymbol{\theta})$ is the marginal likelihood, which is a normalizing constant.

Computing the posterior PDF poses challenges, particularly in formulating the likelihood function $p(\boldsymbol{D}|\boldsymbol{\theta})$ and computing the marginal likelihood $p(\boldsymbol{D})$. For the likelihood function $p(\boldsymbol{D}|\boldsymbol{\theta})$, if the relationship between model parameters and the system responses is explicitly defined (e.g., classical structural dynamics equations

relating the structural stiffness to acceleration responses), it can be easily obtained. However, in the context of inferring discrete states, there is no longer an explicit mathematical model to characterize the relationship between discrete states and system responses. As a result, constructing an analytical likelihood function becomes particularly challenging. For the marginal likelihood $p(\boldsymbol{D})$, the computation is confronted with combinatorial explosion. Since the structure with numerous components typically results in a large number of discrete state parameters, the summation of marginal likelihood is often intractable, i.e.,

$$p(\boldsymbol{D}) = \sum_{\boldsymbol{\theta}} p(\boldsymbol{D}|\boldsymbol{\theta})p(\boldsymbol{\theta}) = \sum\nolimits_{\theta_1} \cdots \sum\nolimits_{\theta_n} p(\boldsymbol{D}|\theta_1, \theta_2, \dots, \theta_n)p(\theta_1, \theta_2, \dots, \theta_n) \tag{2}$$

Furthermore, if we want to know the marginal distribution of each component, the marginalization also needs a complicate accumulation operation:

$$\begin{aligned} p(\theta_i|\boldsymbol{D}) &= \sum\nolimits_{\theta_1} \cdots \sum\nolimits_{\theta_{i-1}} \sum\nolimits_{\theta_{i+1}} \cdots \sum\nolimits_{\theta_n} p(\theta_1, \theta_2, \dots, \theta_n|\boldsymbol{D}) \\ &= \frac{\sum_{\theta_1} \cdots \sum_{\theta_{i-1}} \sum_{\theta_{i+1}} \cdots \sum_{\theta_n} p(\theta_1, \theta_2, \dots, \theta_n, \boldsymbol{D})}{\sum_{\theta_1} \cdots \sum_{\theta_n} p(\boldsymbol{D}|\theta_1, \theta_2, \dots, \theta_n)p(\theta_1, \theta_2, \dots, \theta_n)} \end{aligned} \tag{3}$$

One more challenge lies in the fact that specifying the covariance matrix in the prior distribution to account for the dependencies between model parameters is not suitable for discrete variables. Covariance intrinsically measures the average tendency of two variables to vary together in the same or opposite directions relative to their respective means. Although it can, to some extent, indicate a directional association between discrete variables (e.g., adjacent components are more likely to be in the same or similar states). However, most nominal variables are often subjectively assigned (e.g., component state can be encoded as 1, 2 or any number.), which makes it hard to directly interpret the covariance values as indicators of dependency strength. Moreover, covariance is often suboptimal for characterizing the associations between component states, because adjacent component states often exhibit a complex relationship, which means that when the state of one component changes, the resulting changes in its adjacent components cannot be described by a simple linear function.

To overcome the challenges mentioned before, we employ PGMs to provide a framework for modeling dependencies between component states, linking monitoring data to component states, and enabling efficient posterior estimation.

## 2.2 Bayesian inversion by PGMs

PGMs provide a principled framework for representing complex joint probability distributions over multiple random variables by encoding conditional dependencies in a graph structure. PGMs are broadly categorized into two main types: directed graphical models (Bayesian networks), which represent causal or directional relationships using directed acyclic graphs; and undirected graphical models (Markov Random Fields, MRFs), which capture symmetric, non-directional dependencies through undirected edges. The modeling tool we employ is the pairwise Markov network, which is a subclass of Markov networks.

Specifically, a pairwise Markov network over a graph $\mathcal{G}(\boldsymbol{\nu}, \boldsymbol{\varepsilon})$ is parameterized by a set of potential functions that are defined over cliques [21]. A clique is defined as a subset of nodes that are fully connected to each other, and it serves as the fundamental unit for defining potential functions that encode local dependencies within the probability distribution. To better describe the structural system's behavior, we define potential functions $\psi_{ij}(\theta_i, \theta_j)$ over the clique of adjacent component states $(\theta_i, \theta_j) \in \boldsymbol{\varepsilon}$ to encode the dependencies between components, and $\phi_i(\theta_i, d_i)$ are defined to represent the relationship between each component state and the data measured at the corresponding location. Then the joint distribution can be expressed as:

$$p(\boldsymbol{\theta}, \boldsymbol{D}) = \frac{1}{Z} \prod_{(i,j)\in\boldsymbol{\varepsilon}} \psi_{ij}(\theta_i, \theta_j) \prod_{i\in\boldsymbol{\nu}} \phi_i(\theta_i, d_i) \tag{4}$$

where $\boldsymbol{\nu} = \{\nu_i | i = 1,2, \dots, n\}$ and $\boldsymbol{\varepsilon} = \{(i,j) | i,j \in \boldsymbol{\nu}, i \neq j\}$ represents the nodes and the edges in graph $\mathcal{G}$, respectively; $\theta_i$ denotes the state variable at node $\nu_i$; $d_i$ is the observed data corresponding to the location of the $i$th component; $n$ is the number of nodes in the graph (also called the order of the graph); $Z$ is the partition function that ensures the normalization of the joint probability distribution. For simplicity, here we consider each component state $\theta_i$ as a binary variable, with values of $-1$ or $1$ representing the intact and damaged states, respectively. The number of data points is also assumed to be the same as components.

The potential functions are both taken in the logarithmic form, i.e., $\psi_{ij}(\theta_i, \theta_j) = \exp(-\omega_{ij}\theta_i\theta_j)$ and $\phi_i(\theta_i, d_i) = \exp(-k_i\theta_i d_i)$. Then the distributions are represented as:

$$p(\boldsymbol{\theta}, \boldsymbol{D}) = \frac{1}{Z} e^{E(\boldsymbol{\theta}, \boldsymbol{D})} \tag{5}$$

$$E(\boldsymbol{\theta}, \boldsymbol{D}) = -\sum_{(i,j)\in\varepsilon} \omega_{ij}\theta_i\theta_j - \sum_{i\in\nu} k_i\theta_i d_i \tag{6}$$

where $E(\boldsymbol{\theta}, \boldsymbol{D})$ is the energy of the system with interacting elements. This modeling approach is capable of expressing the energy levels of the node pair under different state configurations, and match their probabilities based on the principle of system energy minimization [21]. The parameters $\omega_{ij}$ and $k_i$, which are to be learned, can also represent the dependency strength between adjacent states.

Consider that data $\boldsymbol{D}$ is observable, it can be regarded as a deterministic part, so the distribution of $\boldsymbol{\theta}$ conditioned on data $\boldsymbol{D}$ can be expressed by:

$$p(\boldsymbol{\theta}|\boldsymbol{D}) = \frac{p(\boldsymbol{\theta}, \boldsymbol{D})}{p(\boldsymbol{D})} = \frac{\frac{1}{Z}\exp\left(-\sum_{(i,j)\in\varepsilon}\omega_{ij}\theta_i\theta_j - \sum_{i\in\nu}b_i\theta_i\right)}{\frac{1}{Z}\sum_{\boldsymbol{\theta}}\exp\left(-\sum_{(i,j)\in\varepsilon}\omega_{ij}\theta_i\theta_j - \sum_{i\in\nu}b_i\theta_i\right)}$$

$$= \frac{1}{Z}\exp\left(-\sum_{(i,j)\in\varepsilon}\omega_{ij}\theta_i\theta_j - \sum_{i\in\nu}b_i\theta_i\right) \tag{7}$$

$$b_i = k_i d_i \tag{8}$$

where $\omega_{ij}$ and $b_i$ denotes edge potential and node potential, respectively. This model is also known as Ising model [39], which was first introduced in statistical physics to represent the energy of a system with interacting atoms.

Actually, through the Markov network modeling for component states and structural responses, the posterior distribution in Bayesian inference is surrogated by the conditional distribution in Eq.(7). We proved that the posterior distribution obtained by the graphical model is identical to the posterior PDF obtained through Bayesian inference. The details of the proof are provided in the **Appendix A**. A simple proof for the equivalence.

Obtaining the marginal distribution of component states requires only determining $\omega_{ij}$ and $b_i$, and then executing inference algorithms on the established model. In this way, there is no need to construct an analytical likelihood function, and the high-dimensional summation is also not required.

### 2.3 Learning of PGMs

The inference of PGMs needs the determination of both model structure and model parameters. The structure of the model indicates on which pairs of state variables the potential function is defined, while the parameters $\omega_{ij}$ and $b_i$ show how these adjacent variables are related and how the monitoring data reflect the component states. However, the incompleteness of structural monitoring data makes it difficult for traditional maximum likelihood estimation (MLE) methods to effectively implement

PGM learning works. Therefore, this study adopts a heuristic approach to determine the model structure and uses the mutual-information-based method for parameter learning.

### 2.3.1 Determination of the PGM structure

Structural learning in PGMs seeks to discover conditional dependencies between variables through measurements, i.e., to actually identify which variables are interrelated. This study uses a heuristic method for structural learning that establishes direct correspondence between the structure of graphical model and structure's geometry. Specifically, the structure of the graphical model is equivalent to the spatial configuration of the considered components. An example is shown in Figure 1. This approach maintains physical consistency with structural system behavior through intuitive alignment, and it is also a direct utilization of structural topology prior. From a mechanical perspective, there are complex force transmission paths and interaction mechanisms between structural components. An alteration in the mechanical behavior of one component propagates to adjacent components through connections. This heuristic method can effectively capture direct correlation patterns between neighboring components through edge connections.

### 2.3.2 Learning potential function $\boldsymbol{\phi_i(\theta_i, d_i)}$

When the relationship between the discrete model parameters $\boldsymbol{\theta}$ and the measurement $\boldsymbol{D}$ is complex, learning this part is quite challenging. To mitigate the effects of noisy raw data, the feature data which is sensitive to anomalies is employed (e.g., modal data). Consider that the measurement $d_i$ is not a discrete variable, we can learn the potential function $\phi_i(\theta_i, d_i)$ across all possible realizations of $\theta_i$ to comprehensively account for all scenarios. According to the assumption that $\theta_i$ takes the value of -1 and 1 for respectively representing intact state and damaged state, there are two scenarios to be considered. The first scenario is that the structure is intact. Since we generally have monitoring data for intact structures, we can use a Gaussian mixture model (GMM) based density estimation method to obtain $\phi_i(\theta_i = -1, d_i)$. The probability distribution of $d_i$ can be expressed as:

$$p(d_i) = \sum_{k=1}^{K} \pi_k \mathcal{N}(d_i; \mu_k, \Sigma_k) \tag{9}$$

where $K$ is the number of Gaussian models; $\pi_k$ is the weight of the $k$th Gaussian distribution; and $\mu_k$ and $\sum_k$ denote the mean and covariance of the Gaussian distribution, respectively. To determine the appropriate number of mixture components $K$, we employ the Bayesian Information Criterion (BIC) for model selection [40], which balances model fit and complexity to avoid overfitting. Specifically, we train multiple GMMs with candidate values of $K$ (e.g., $K = 1$ to $K = 10$), and select the model that minimizes the BIC value. This data-driven approach ensures that the chosen $K$ captures the underlying distribution without unnecessary complexity, enhancing the robustness of the density estimation in practical scenarios where the true distribution may be multimodal. Based on the Expectation-Maximization (EM) algorithm [41], $p(d_i)$ can be learned with the optimal $K$.

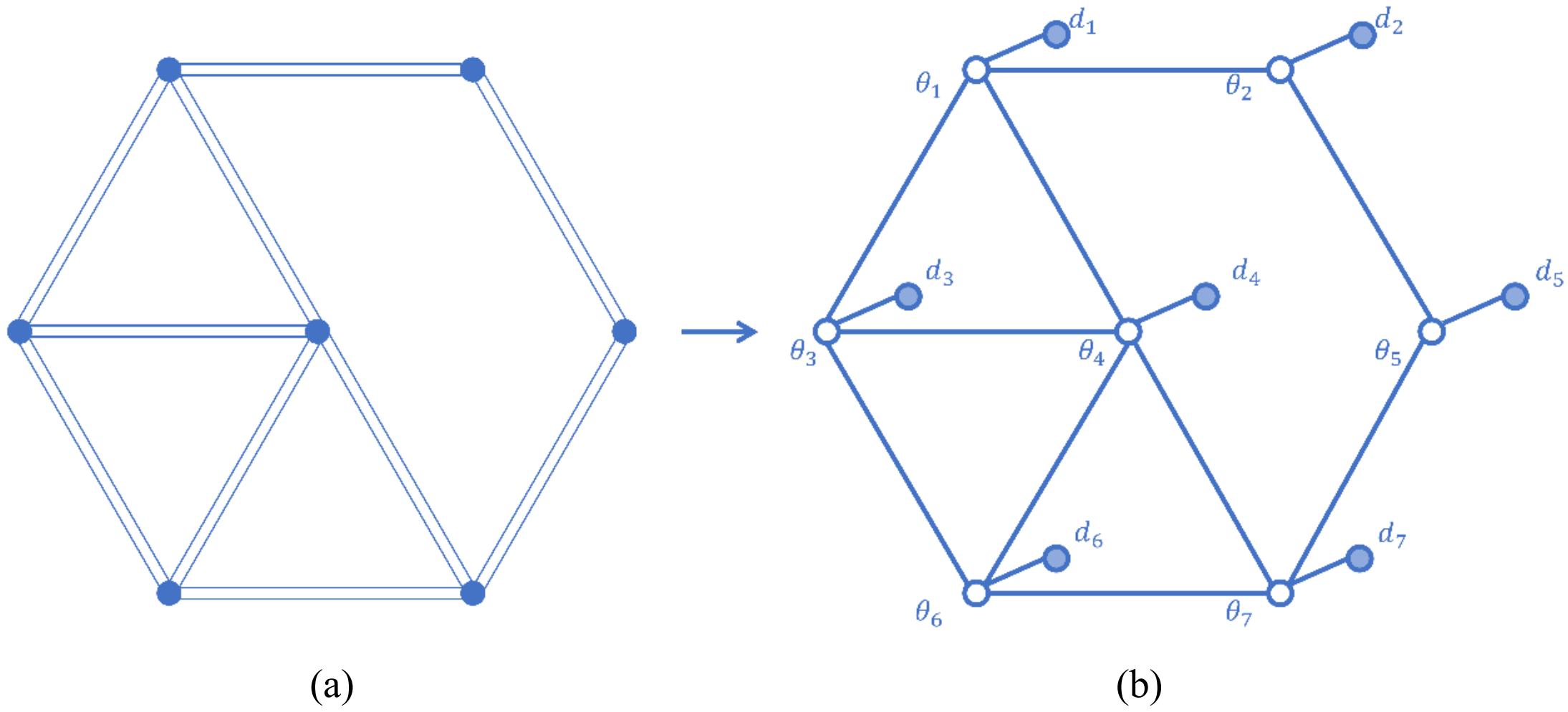

Figure 1. An example of the structure learning method. (a) a structure with sensors. (b) the probabilistic graphical model corresponding to (a).

The second scenario involves learning $\phi_i(\theta_i = 1, d_i)$. If feature data for the current structure (or the identical structure) in a damaged state is available, GMM can still be employed for density estimation. However, in many cases, estimating this distribution is challenging due to the lack of damage data. Under such constraints, the maximum entropy principle [42] provides a theoretical foundation for $p(d_i)$ to adopt a uniform distribution over the feature data domain. The data domain for $d_i$ is defined as the plausible range of the feature data, and its boundary needs to be determined before obtaining the target distribution $p(d_i)$.

Some features may have a clearly defined domain (e.g., the feature in [43]), in which case it can be directly adopted as the data domain boundary. If the domain is undefined, the preferred approach is to determine the data domain using a structural finite element model or numerical simulations. By simulating different damage scenarios to generate response data under damaged conditions, the reasonable variation range of feature data $d_i$ can be statistically derived. In addition, if the current conditions are not sufficient to support the calculation of the physical model, it is recommended to use a data-driven approach based on the statistics of the intact-state data. Specifically, a set of measurements of intact-state structure can be used to calculate their mean $\mu$ and standard deviation $\sigma$, and then the data domain can be defined as $[\mu - 3\sigma, \mu + 3\sigma]$, which follows the empirical "3-sigma rule" to cover the vast majority of potential changes. This range can be further adjusted if more information is known. Furthermore, expert knowledge and empirical understanding of structural dynamics can also be used for informing the choice of the distribution.

After obtaining the distribution of $p(d_i)$, the parameters $b_i$ can be estimated by:

$$\begin{cases} p_{\theta_i=-1,d_i}\left(\overline{d}_i\right) = P_{u_i} = \exp(-b_i + c_i) \\ p_{\theta_i=1,d_i}\left(\overline{d}_i\right) = P_{d_i} = \exp(b_i + c_i) \end{cases} \tag{10}$$

$$b_i = \frac{1}{2}\left(ln(P_d) - ln(P_u)\right) \tag{11}$$

where $p_{\theta_i=-1,d_i}$ and $p_{\theta_i=1,d_i}$ denote the PDF of data given different component state; $\overline{d}_i$ represents the mean value of feature data in repeat tests; $P_{u_i}$ and $P_{d_i}$ denote the probability density of $\overline{d}_i$ in different distributions; and $c_i$ is a constant that does not affect the results.

### 2.3.3 Learning potential function $\psi_{ij}(\theta_i, \theta_j)$

The estimation of $\psi_{ij}(\theta_i, \theta_j)$ is an intractable problem due to the unavailability of data in numerous SHM scenarios [21]. As an alternative, we use mutual information to approximately learn this parameter [26]. According to the definitions in **Section 2.2**, the structural component state variables $\theta_i$ take the values of -1 and 1 for representing an intact state and a damaged state, respectively. An example for a simple case is given below that can be generalized to complex models. Considering a pairwise Markov network with two nodes $v_1$ and $v_2$, the potential function is expressed as:

$$\psi_{12}(\theta_1, \theta_2) = \exp(-\omega_{12}\theta_1\theta_2) \tag{12}$$

where the binary variables $\theta_1$ and $\theta_2$ are specified with the value of $-1$ or $1$, so we can derive that

$$\psi_{12}(\theta_1 = 1, \theta_2 = 1) = \psi_{12}(\theta_1 = -1, \theta_2 = -1) = e^{-\omega_{12}} \tag{13}$$

$$\psi_{12}(\theta_1 = -1, \theta_2 = 1) = \psi_{12}(\theta_1 = 1, \theta_2 = -1) = e^{\omega_{12}} \tag{14}$$

This implies that the probability of a node having the same state as its neighbor is:

$$\frac{e^{-\omega_{12}}}{(e^{-\omega_{12}}+e^{\omega_{12}})}=\frac{1}{(1+e^{2\omega_{12}})} \tag{15}$$

By generalizing this to graphical models with more than two nodes, the potential functions can be represented as:

$$\psi_{ij}(\theta_i,\theta_j)=\begin{cases}1-p_{ij} & if\ \ \theta_i\neq\theta_j\\ p_{ij} & if\ \ \theta_i=\theta_j\end{cases},0\leq p_{ij}\leq 1 \tag{16}$$

where $p_{ij}$ denotes the probability that structural states at adjacent two nodes are the same. With such a potential function, the state at one node can be reflected in its neighboring nodes. Due to the fact that the state of each component can be reflected by the response monitored at the corresponding location, it is appropriate to estimate parameter $p_{ij}$ based on the similarity in the distribution of adjacent monitoring data. The mutual information is used to quantify data similarity between nodes $i$ and $j$. Specifically, we define $p_{\theta_i,d_i}(d_i)$ as the PDF of data measured at node $i$. The similarity modeling function is defined as:

$$s(d_i,d_j)=\left(\frac{p_{\theta_i=\theta_j=-1,d_i,d_j}(d_i,d_j)}{p_{\theta_i=-1,d_i}(d_i)p_{\theta_j=-1,d_j}(d_j)}\right) \tag{17}$$

where $p_{\theta_i=\theta_j=-1,d_i,d_j}(d_i,d_j)$ represents the joint probability distribution of measurements with two neighboring nodes $i$ and $j$ both intact. The distributions $p_{\theta_i,d_i}(d_i)$ and $p_{\theta_i=\theta_j=-1,d_i,d_j}(d_i,d_j)$ can be determined by the methods illustrated in the last part. By sampling from the distribution of $p_{\theta_i=\theta_j=-1,d_i,d_j}(d_i,d_j)$, the mutual information between $i$ and $j$ is estimated by:

$$I(d_i,d_j)\approx\frac{1}{N}\sum_{k=1}^{N}log\left(s(d_i^k,d_j^k)\right) \tag{18}$$

where $N$ denotes the number of samples and $d_i^k$ is the $k$th sample. For a more convenient calculation, the normalized mutual information is used, which is defined as:

$$N(d_i,d_j)=\frac{I(d_i,d_j)}{\sqrt{i(d_i)i(d_j)}} \tag{19}$$

where $i(d_i)=-\frac{1}{N}\left(\sum_{k=1}^{N}log\left(p(d_i^k)\right)\right)$. The normalized mutual information directly quantifies dependency between neighboring nodes. Consider the extreme case that node $i$ and its neighbor node $j$ are completely dependent on each other, the state of one node can be determined by the other. In this case, $N(d_i,d_j)$ is identified as 1 and the probability denoted by $p_{ij}$ takes the value of 1 at the same time. Another extreme case is that $i$ and $j$ are independent, and it is obvious that $s(d_i, d_j) = 1$. Due to the properties of the logarithmic function, the value of $N(d_i, d_j)$ is zero, which means that the state of one node cannot provide any information about its neighbor node. In other words, even if the state of a node has been determined, the probability of each state occurring in its neighboring nodes is the same, i.e. $p_{ij} = 0.5$. In general situations, the structure lies between two extreme cases and the probability of regular cases can be estimated by linear interpolation:

$$p_{ij} = 0.5\left(1 + N\left(d_i, d_j\right)\right) \tag{20}$$

Once the parameters $p_{ij}$ are learned, $\omega_{ij}$ could be estimated by the following equations:

$$\begin{cases} \psi_{ij}\left(\theta_i = 1, \theta_j = 1\right) = \psi_{ij}\left(\theta_i = -1, \theta_j = -1\right) = p_{ij} = \exp\left(-\omega_{ij} + c_{ij}\right) \\ \psi_{ij}\left(\theta_i = 1, \theta_j = -1\right) = \psi_{ij}\left(\theta_i = -1, \theta_j = 1\right) = 1 - p_{ij} = \exp\left(\omega_{ij} + c_{ij}\right) \end{cases} \tag{21}$$

$$\omega_{ij} = \frac{1}{2}\left(ln\left(1 - p_{ij}\right) - ln\left(p_{ij}\right)\right) \tag{22}$$

where $c_{ij}$ is a constant that does not affect the results.

To summarize, the proposed method is implemented through the following steps.

(1) Obtaining the monitoring data sets from sensors on the structure.

(2) Computing the parameters $\omega_{ij}$ and $b_i$ by using Eq.(9) ~ Eq.(22). During this phase, prior structural knowledge may be incorporated when available.

(3) Feeding these parameters into GNNs (The details are shown in the next section), and the final inference results will be output.

## 3. Size-generalizable GNN for PGM inference

A straightforward algorithm to make inference of PGMs is variable elimination, which substitutes all possible values of the non-target variables into the joint distribution and sums them up. However, this approach is NP-hard and it is difficult to implement when the graphical models are large (i.e., multiple components problem with high-dimensional parameters) [21]. Approximate inference algorithms, such as belief propagation (BP) and Markov Chain Monte Carlo (MCMC), often prove more practical than exact solutions for general implementations. BP has the ability to achieve accurate results in acyclic graphs. However, it performs poorly in graphs containing cycles, as information propagation becomes non-terminating [27]. MCMC excels in exact inference, but its computational efficiency significantly decreases as the graph scale increases (e.g., the dimension of parameters increases). **Appendix B**. General algorithms for PGMs inference provides a detailed introduction to the VE, BP, and MCMC algorithms. In this study, a GNN-based inference algorithm is employed to deal with both problems, as we illustrate in the following section.

### 3.1 GNN architecture for PGM inference

Graph neural networks [30] (GNNs) were developed to process graph-structured data characterized by complex interdependencies. GNN architectures comprise nodes and edges, where each element carries a vector representation encoding its hidden state. These hidden states undergo iterative updates through nonlinear transformations that integrate information from the node itself, its adjacent neighbors, and features of connecting edges. Once the hidden states converge or the iteration runs to the specified step, another non-linear function will decode the updated hidden state and outputs the results. In general, the functions are performed by neural networks. Their parameters are shared to ensure that the pattern of information aggregation is consistent throughout the entire graph.

A critical challenge in applying GNNs to graphical model inference is designing a mapping scheme between graphical model and GNN. In other words, it is about how to design inputs, outputs, and network operations to adapt for inference tasks. Consider that graphical model parameters $\omega_{ij}$ and $b_i$ critically determine inference outcomes, they serve as natural input candidates. In addition, the vector-valued hidden state defined at each node is considered as key content during the operation, as it may encode the critical probabilistic information for inference. The GNNs ultimately produce a vector output approximating the target marginal distribution. The design of GNN in this study is based on the process of BP algorithm, i.e., the core concept is to propagate information between adjacent nodes. Formally, assume that each node is assigned with a P-dimensional vector $h_i^t$ which denotes the hidden state of the node $i$ at step $t$. Similar to the process of BP, the message passed from node $i$ to node $j$ is defined as:

$$m_{i\to j}^{t+1} = F\left(h_i^t, h_j^t, \omega_{ij}, b_i, b_j\right) \tag{23}$$

where $F$ is the deep neural network whose parameters are shared; and $m_{i\to j}^{t+1}$ is a Q-dimensional vector. The information received at each node can be aggregated as:

$$m_i^{t+1} = \sum_{j\in N(i)} m_{j\to i}^{t+1} \tag{24}$$

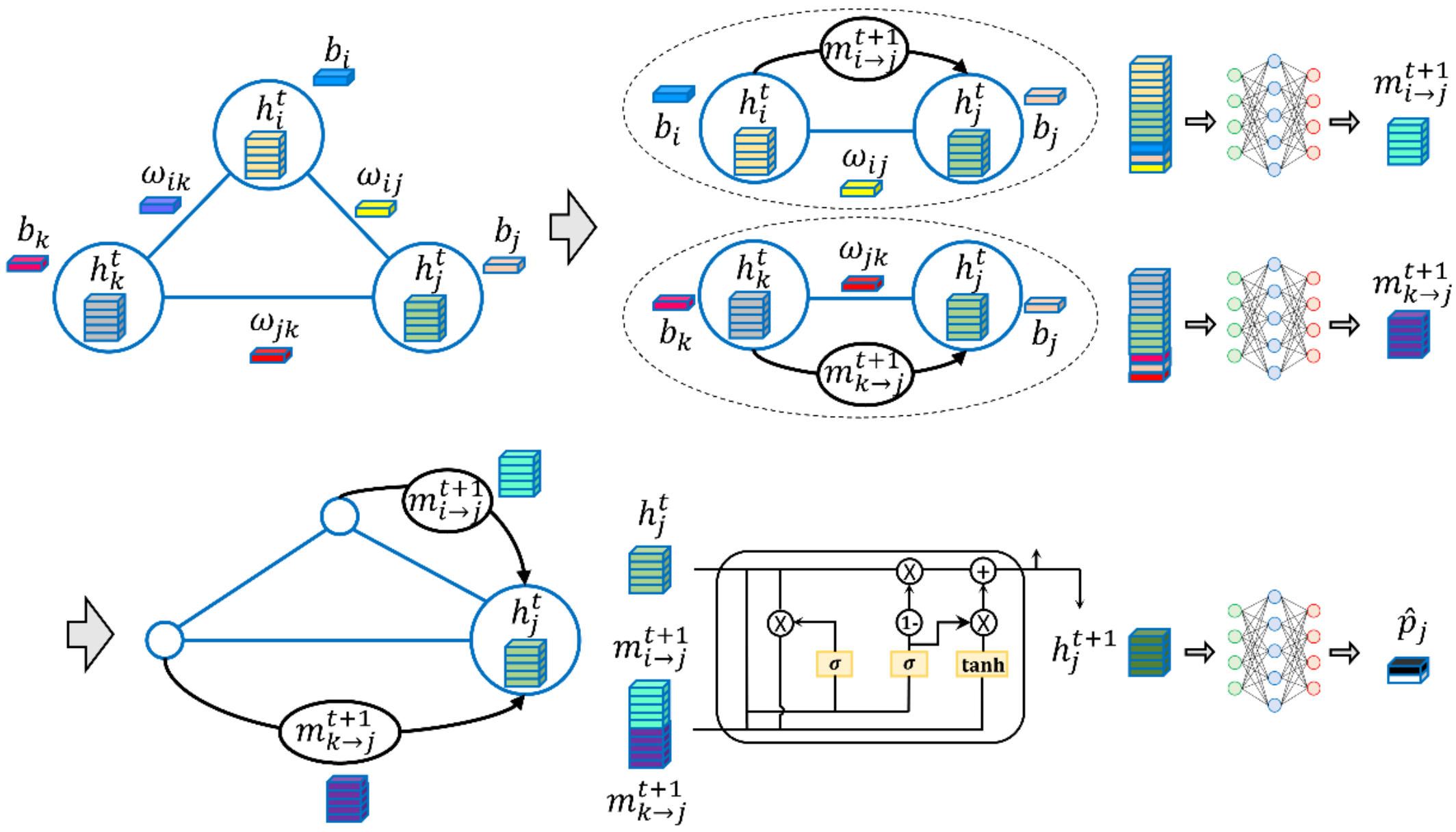

Figure 2. The process of making inference on graphical models by GNNs. This figure shows the process of updating the hidden state at node $j$. The messages coming from adjacent nodes $i$ and $k$ are encoded by the deep neural networks. Then these messages are aggregated at node $j$ and the hidden state at time $t$ is updated by GRUs. At the end of the iteration, the updated hidden state is decoded by another neural network and outputs the target distribution.

where $N(i)$ denotes all the neighboring nodes of $i$.

Next, the hidden state of each node is updated based on the latest information, and this process is implemented by the Gated Recurrent Units [44] (GRUs):

$$h_i^{t+1} = G(h_i^t, m_i^{t+1}) \tag{25}$$

where $G$ represents the recurrent neural network with GRUs. When the iteration reaches the specified step $T$, another function decodes the updated hidden states and outputs the target distribution $\hat{p}$:

$$\hat{p} = R(h_i^T) \tag{26}$$

where $R$ is the deep neural networks; $\hat{p}$ denotes the estimated marginal probabilities. Figure 2 shows the entire process of using GNNs for graphical model inference.

### 3.2 Size generalization of GNNs

GNNs require rigorous training schemes to achieve accurate probabilistic inference. Training labels typically require exact inference algorithm (such as VE) for accuracy, but the computational complexity of such algorithm grows exponentially with the increase of model parameter dimensions. As a result, obtaining a well-trained model consumes substantial time and computational resources in practical applications [25].

Many problems could be addressed efficiently if GNNs can learn the intrinsic properties of inference on small-scale graphical models and then successfully apply it to large-scale graphs. However, there is currently no universal theoretical framework in the field of GNN size generalization to guide the training of GNNs. Notably, within the domain of in-distribution generalization for MPNNs, theoretical frameworks for generalization boundary analysis have been established under varying assumptions and architectures [45]. While these investigations do not provide direct methodological instructions for network training, their analytical frameworks systematically identify critical factors influencing the generalization capacity of MPNNs. This foundation enables exploration of network training strategies which enhance MPNNs' size generalization capabilities. For example, the Rademacher complexity analysis of MPNNs for in-distribution generalization can be formulated through layer-wise decomposition of hypothesis space complexity. Let $\hat{\mathcal{R}}(\mathcal{T}_\gamma)$ denotes the empirical Rademacher complexity of the hypothesis class $\mathcal{T}_\gamma$, if the maximum degree for any node is at most $\tilde{d}$, the empirical Rademacher complexity $\hat{\mathcal{R}}(\mathcal{T}_\gamma)$ is bounded by:

$$\hat{\mathcal{R}}(\mathcal{T}_\gamma) \leq \frac{4}{\gamma N} + \frac{24 d \mathcal{B}_\beta Z}{\gamma \sqrt{N}} \sqrt{3 \log Q} \tag{27}$$

$$Q := 24 B_\beta \sqrt{N} \max\{Z, \mathcal{M} \sqrt{d} \max\{\mathcal{B}_h \mathcal{B}_1, \overline{R} \mathcal{B}_2\}\} \tag{28}$$

where $N$ denotes the number of training samples; $d$ is the dimension of node feature space. The details of this theory can be found in [45].

In graph theory, the degree of a node is defined as the number of edges incident to this node, and an $n$-order graph represents a graph with $n$ vertices. Considering that the degree, order, and number of training samples can have an impact on the generalization boundary of GNNs, in this work, several properties of the graph (average unique node degree, graph order, and the number of training samples) are studied to inform how to effectively train the GNNs and generalize them to make accurate and efficient inference on larger graphical models. The average unique node degree in a graph is defined as the average value of the unique node degrees, where 'unique' represents that the same degree in the graph is only counted once. The analysis of these graph properties on the size generalization performance of GNNs is elaborated in the next section. More details about the training of GNNs are also provided in **Section 4**.

## 4. Synthetic data study

In practical engineering applications, graphical models for engineering structures are often high-dimensional, as most of the structures are typically composed of multiple components. Training GNNs on models with the same size leads to a high cost in label acquisition. In order to solve this problem, we intend to train the GNNs on small graphical models and generalize them to larger ones. This section discusses how several graph properties affect the size generalization performance of GNNs in the context of PGM inference, and summarizes how to apply them to make inferences on large graphical models, thereby handling the computation challenge in high-dimensional problems.

### 4.1 Parameter settings of GNN training

The basic settings of training the GNNs are introduced as follows. The parameters of training graphs are sampled from different Gaussian distributions, where $\omega_{ij} = \omega_{ij} \sim \mathcal{N}(0,1)$ and $b_i = b_i \sim \mathcal{N}(0, 0.25^2)$. Hidden vectors and message vectors in GNNs are set to 5-dimensional vectors, while the output layer produces a 2-dimensional vector indicating probabilities of two different states. The hidden state vectors are iterated 10 times before being output, i.e., T=10 in Eq. 26. The activation function in the neural network $F$ is specified by a rectified linear unit (ReLU), while sigmoid function $\sigma(x) = 1/(1 + \exp(-x))$ is taken as the activation function for the output network $R$. In addition, both networks have two hidden layers and each layer contains 64 units. GNNs are trained via supervised learning, optimized by minimizing the cross-entropy loss:

$$L(p_i, \widehat{p_i}) = -\sum_i p_i(x_i) \log \widehat{p_i}(x_i) \tag{29}$$

where $p_i$ is the real probabilities and $\widehat{p_i}$ is the estimated probabilities. The GNNs are trained using the ADAM optimizer [46]. The learning rate and training epochs are set to 0.001 and 100, respectively.

### 4.2 The effects of graph properties on GNN size generalization performance

To analyze how graph properties affect size generalization performance, we trained multiple GNN models using graphs with varying graph properties. Our dataset includes graphs with orders ranging from 4 to 100 and average unique node degrees ranging from 2 to 9 (the properties of graphs with fewer than 3 nodes are basically the same, so we did not include them in the dataset). Some instances of the training and testing graphs are shown in Figure 3. We used Kullback-Leibler (K-L) divergence between predicted distribution and ground truth as the accuracy metric, where lower values correspond to higher estimation accuracy. The performances of three algorithms (GNN, BP and MCMC) are also compared.

### 4.2.1 Training set size

Firstly, we attempted to determine the optimal training set size by observing the inference performance under different graph orders and degrees. The results are shown in Figure 4. It can be seen that when the average unique node degree is fixed, even if the graph order changes, the training samples required to achieve good performance of GNN do not change significantly (all around 1000). Instead, when the graph order is fixed, as the average unique node degree increases, the number of training samples required to achieve better results will also increase accordingly.

### 4.2.2 Average unique node degree

The next experiment is implemented to investigate the impact of the average unique node degree of the training graphs on the size generalization performance of GNNs. We trained several GNN models on graphical models with average unique node degree from 2 to 9, and the number of training samples for each degree scope depends on the results in Figure 4. The testing results of the degree experiment are shown in Figure 5. The horizontal and vertical axes in the figure represent the average unique node degree of the training and testing sets, respectively. Color intensity in the figure corresponds to K-L divergence values, with darker hues indicating lower divergence (i.e., higher accuracy). The subfigure (a) demonstrates the inference performance of the GNNs on graph order-invariant graphical models (the graph order of the training and testing sets is fixed at 10), and (b), (c), and (d) shows the inference performance of the GNNs on larger graphical models (16, 36 and 100, respectively). It can be seen in this figure that GNNs achieve better performance on test graphical models when the average node degree is similar to that of the training set. When training and testing graphs share identical orders, the performance improvement becomes prominent. As the graph order in the testing set increases, even though there is some degradation, it can still be seen that this rule is satisfied.

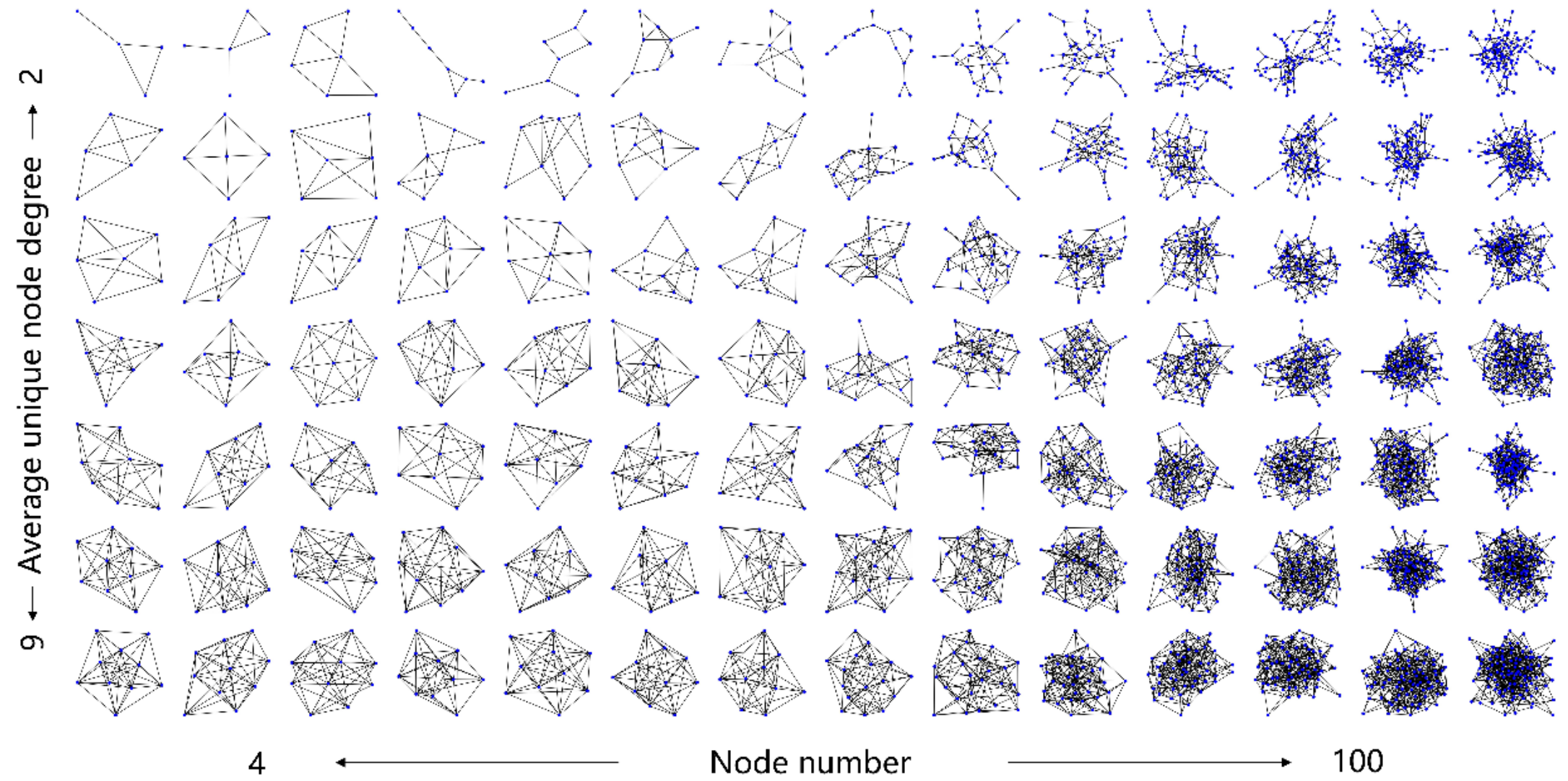

Figure 3. Some instances of the training and testing sets. Our dataset includes graphs with node counts ranging from 4 to 100, with an average unique node degree spanning from 2 to 9. The connections between nodes are randomly generated, and the number of connections depends on the pre-set average unique node degree. (Note that when the average unique node degree is $k$, a graph that can meet the requirements must have at least $k$+1 nodes).

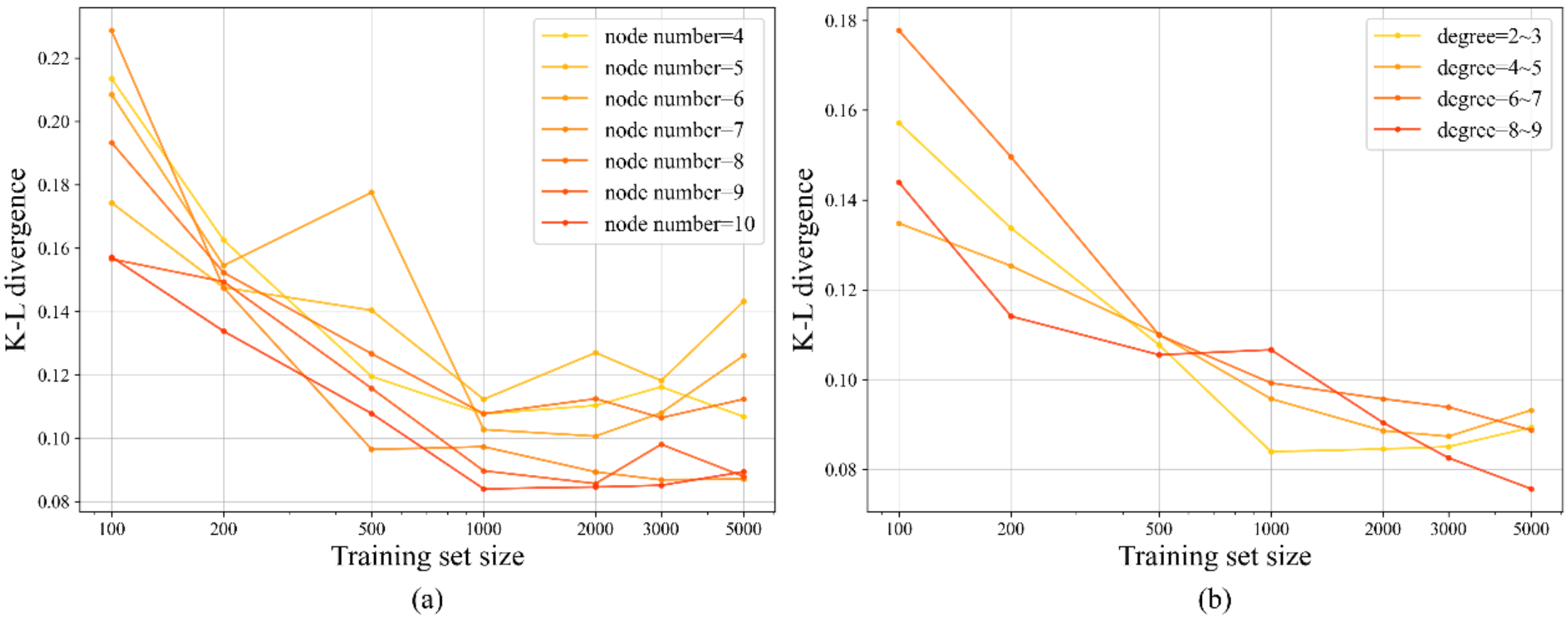

Figure 4. The experiment on finding the optimal training set size. The subfigure (a) shows the case where the average unique node degree is fixed at 2~3, and the graph order of the training graphs ranges from 4 to 10. The subfigure (b) shows the case where the graph order of the training graphs is fixed at 10, and the average unique node degree ranges from 2 to 9.

Consequently, it can be concluded from the experiment results that the average unique node degree is a major factor affecting the size generalization performance of GNNs, i.e., when the average unique node degree of the graphs used for training is the same or close to that of the target graphical model, inference tasks on larger graphical models can be more accurately implemented. In the context of structural health monitoring applications, based on the structure learning method in this paper, the graph properties of graphical models established for common structures are often partly similar. The average unique node degree of these graphical models often falls within a suitable range, as the adjacent nodes connected to each node of the structure are usually not too many (to save materials and reduce weight) nor too few (to ensure stability and safety). For example, the graphical model (shown in Figure 6) corresponding to the IASC-ASCE benchmark model [47] has the unique degree from 4 to 7, and its average unique node degree ranges from 5 to 6. Similarly, for common civil engineering structures, if the structure learning method described in this article is used to establish the graphical structure, the average unique node degree usually does not deviate significantly from the graph structure in the example. This facilitates GNN training by enabling efficient model configuration based on graph properties.

### 4.2.3 Graph order in the training set

To explore the practicality of our method, we implemented another experiment to investigate the size generalization performance of GNNs at the specific average unique node degree. The results are shown in Figure 7. On the basis of the experiment results in the previous section, we tested the size generalization ability of GNNs while making sure that the average unique node degree of the training and testing sets was the same. The subfigure (a) shows the inference performance of the GNN on seen graphs. The comparison of GNNs with traditional algorithms BP as well as MCMC is also displayed in (a). Subfigures (b), (c), and (d) illustrate size generalization cases with average unique node degrees of 2-3, 3-4, and 5-6, respectively. Obviously, as the graph order in the training set increases, the inference accuracy on unseen large graphical models becomes higher. Therefore, by determining the various properties of the target graphical model to be inferred, one can choose a GNN training method that can ensure computational accuracy while spending less computational resources. In order to display the performance of inference more intuitively, Figure 8 shows a scatter plot of the inference results when the average unique node degree is fixed at 2-3 and the training graph order is fixed at 9. The horizontal and vertical axes represent the true results and GNN predicted results, respectively. Proximity of data points to the diagonal line indicates prediction accuracy. Combined with the results in Figure 7, it can be concluded that when the K-L divergence is less than 0.15, the inference is relatively accurate. The inference performance of GNNs is also compared with algorithms (BP and MCMC) in Figure 8, and four metrics [15], including the coefficient of determination (R2), mean absolute error (MAE), mean squared error (MSE) and root mean squared error (RMSE), are provided to quantitatively assess the performance of three algorithms. The results in Table 1 indicate that under almost all conditions, GNN performs better than BP and MCMC.

To leverage the size generalization capacity of GNNs in engineering applications, the following steps are recommended: First, establish a graphical model representation of the target structure. Then, design a GNN training strategy by analyzing the structural properties of the target graph structure. Subsequently, the graphical models used for GNN training need to be selected. When the training graphs and the target graph structure have identical average unique node degrees, the minimum training graph order which satisfies the predefined inference accuracy should be selected. This methodology can achieve computational efficiency optimization without compromising prediction reliability. Additionally, it is worth noting that if the graph to be inferred is too large, increasing hyper-parameters such as T in Eq.26 is also an effective approach. This action can expand the message propagation range and effectively improve accuracy, though at the expense of additional computational cost.

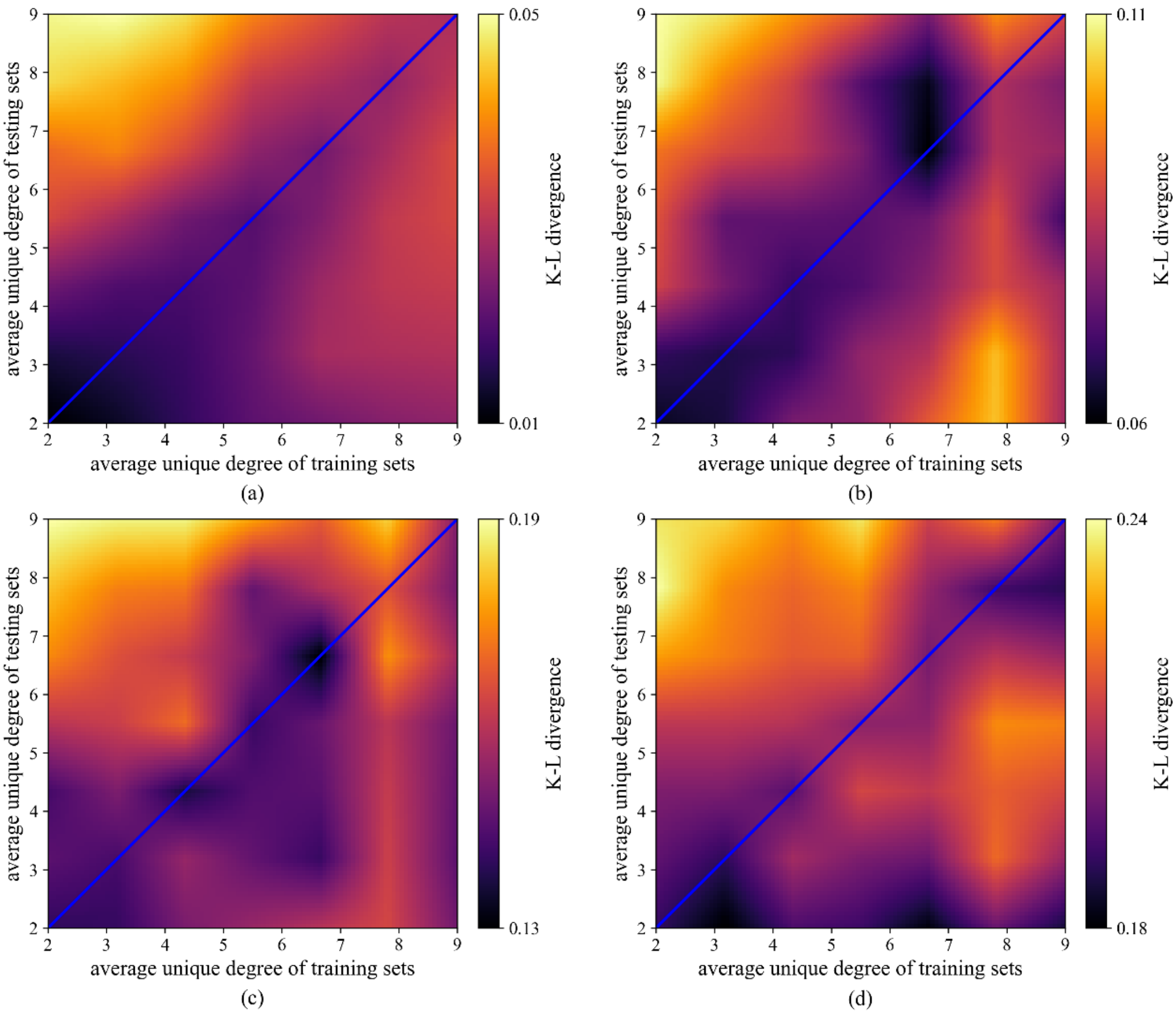

Figure 5. The experiment results for the analysis of the influence of average unique node degree on size generalization performance. The graph order of the graphical models in (a), (b), (c) and (d) used to test the GNNs performance is set to 10, 16, 36 and 100, respectively. The horizontal and vertical axes represent the average unique node degree of the training and testing sets, respectively. The colors in the figure represent their K-L divergence values, with smaller values indicating higher inference accuracy.

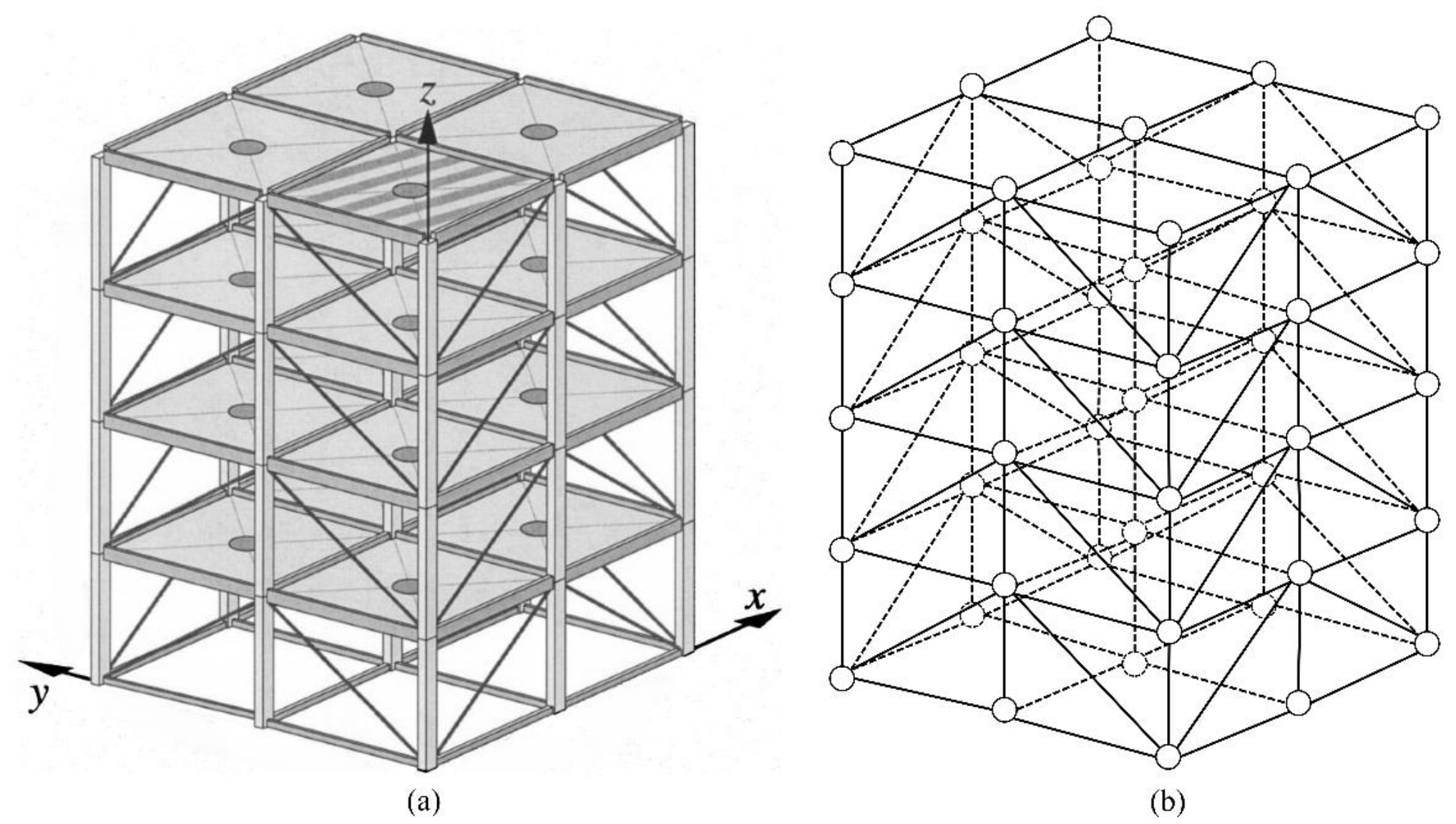

Figure 6. The structure of the IASC-ASCE benchmark (a) and its corresponding graphical model (b).

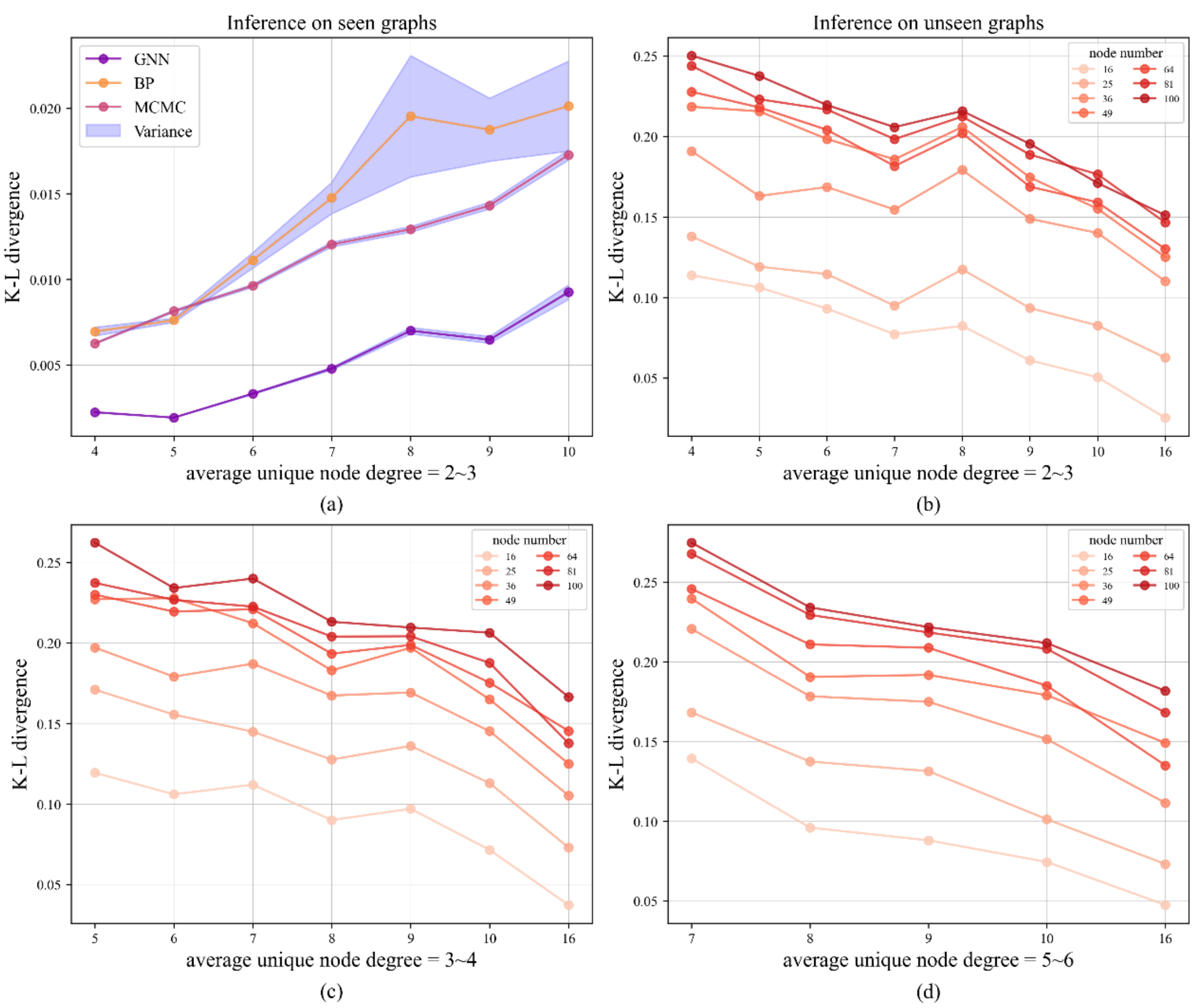

Figure 7. The experiment to investigate the size generalization performance of GNNs at the specific average unique node degree. The horizontal axis represents the graph order of the graphical models used to train the GNNs. The vertical axis represents K-L divergence as usual.

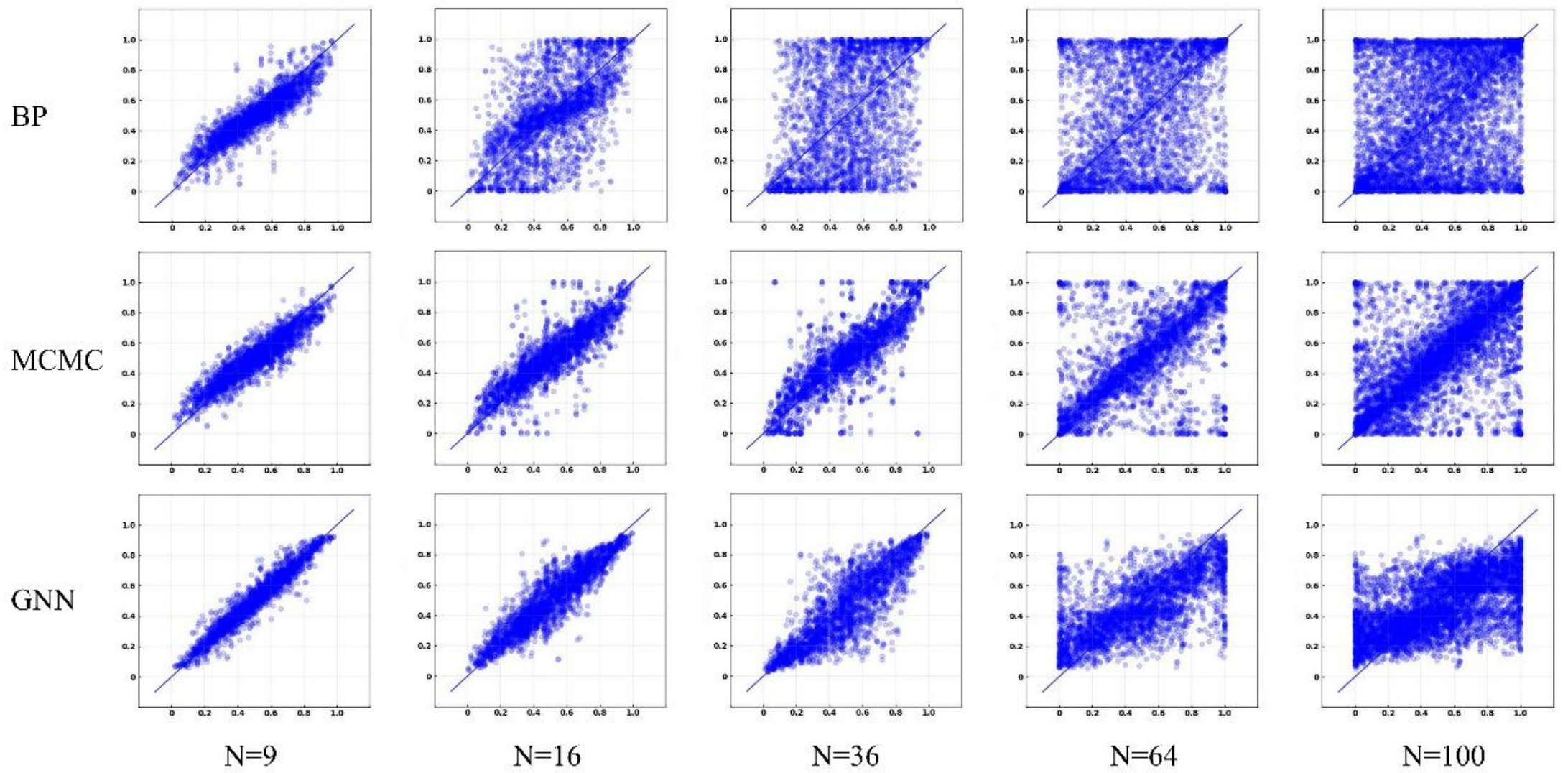

Figure 8. The scatter plot of the inference results in the size generalization experiment. The average unique node degree is fixed at 2~3, and the training graph order is set to 9. The horizontal and vertical axes represent the true results and predicted results, respectively. The closer the points in the graph are to the diagonal line, the more accurate the results are. The performance of three algorithms are compared.

| Metric | | $R^2$ | | | MAE | | | MSE | | | RMSE | | |
|---|---|---|---|---|---|---|---|---|---|---|---|---|---|
| Algorithm | | BP | MCMC | GNN | BP | MCMC | GNN | BP | MCMC | GNN | BP | MCMC | GNN |
| N | 9 | 0.8022 | 0.8235 | **0.9349** | 0.0645 | 0.0635 | **0.0376** | 0.0074 | 0.0066 | **0.0024** | 0.0862 | 0.0815 | **0.0495** |
| | 16 | 0.1365 | 0.8026 | **0.8776** | 0.1575 | 0.0772 | **0.0625** | 0.0458 | 0.0105 | **0.0065** | 0.2140 | 0.1023 | **0.0806** |
| | 36 | -0.4361 | 0.6204 | **0.6605** | 0.2491 | **0.1174** | 0.1234 | 0.1089 | 0.0288 | **0.0257** | 0.3300 | 0.1696 | **0.1604** |
| | 64 | -0.5334 | 0.2531 | **0.3798** | 0.2793 | 0.1567 | **0.1885** | 0.1482 | 0.0722 | **0.0599** | 0.3850 | 0.2687 | **0.2448** |
| | 100 | -0.3848 | -0.4055 | **0.2693** | 0.2819 | 0.2581 | **0.2412** | 0.1694 | 0.1720 | **0.0894** | 0.4116 | 0.4147 | **0.2990** |

Table 1. Comparison of quantitative metrics for inference accuracy of three algorithms. The average unique node degree of training and testing graphs is fixed at 2~3, and the training graph order is set to 9.

### 4.2.4 Inference efficiency comparisons

A comparative assessment of computational time expenditures across three algorithms was conducted in Figure 9. The GNN architecture demonstrated significantly enhanced computational efficiency compared to conventional methods, particularly when performing inference operations on large-scale graph structures. Notably, variations in average unique node degree exerted measurable influence solely on BP algorithm performance – an observation aligning with fundamental algorithmic mechanisms. This difference in sensitivity highlights GNNs' structural scalability advantages in complex network analysis tasks.

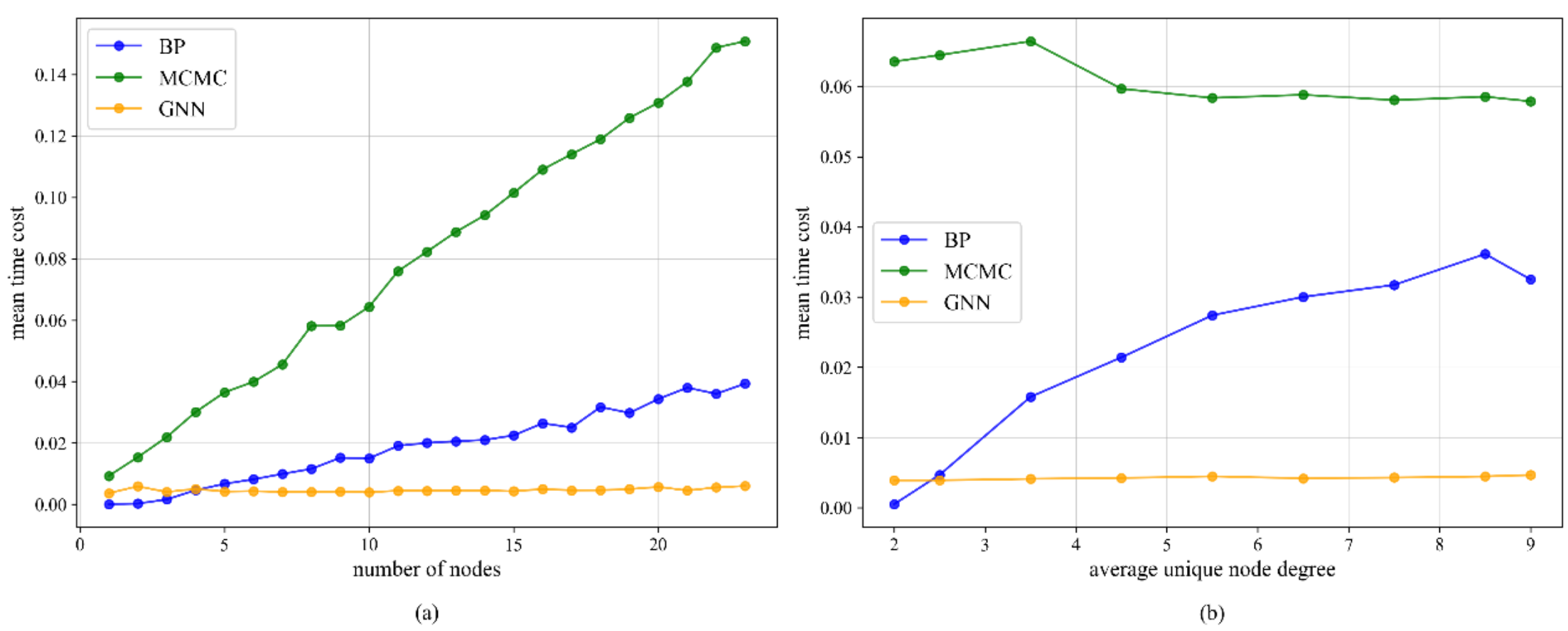

Figure 9. Comparison of computational efficiency among three algorithms. The subfigure (a) shows the trend of computation time as the graph order increases and the subfigure (b) shows the trend of computation time as the average unique node degree increases.

## 5. Experiments: inferring discrete states of a synthetic truss structure

In this part, we validate the proposed method based on some experiments on a truss bridge, which consists of 160 bars with annular cross-sections. The longitudinal length of this structure is 5.5118m, with 14 bars of 0.3937m each. The transverse length is also 0.3937m, while the vertical height is 0.40m. All elements share identical annular cross-sections with outer and inner diameters of 0.01554 m and 0.01087 m, respectively. The materials used to construct the structure have a Young's modulus of 18.5GPa and a shear modulus of 79.3GPa, while the density is 8000kg/m3. The image of the synthetic truss structure, its member layout, the cross-section of the components, and their sizes are shown in Figure 10. Three-dimensional band-limited white noise excitations are applied at all joint locations on the structure, with vertical excitation being dominant.

Our experiment consists of one case that the structure is intact and four cases with different damages. Each damage case is simulated by a different stiffness loss at a different bar of the truss structure. The dataset used for validation was obtained from vibration videos capturing the structure's vibration in both its intact and damaged states. Figure 11 shows two example frames from these videos. The displacements of the joints in the front row of the truss are extracted from the videos. The pixel at each joint is considered as a pseudo-sensor, and the structure of the graphical models is learned as shown in Figure 12. The specific damage location and corresponding stiffness reduction for each case are as follows: In Case 1, the element connecting nodes 3 and 4 experienced a stiffness loss of 52.7 %; in Case 2, the element between nodes 4 and 12 had a stiffness reduction of 52.7 %; in Case 3, the element linking nodes 6 and 7 suffered a stiffness loss of 72.5 %; and in Case 4, the element connecting nodes 2 and 10 was subjected to a stiffness reduction of 72.5 %. In order to avoid the impact from external loads, we use the fundamental mode of structural vibration to infer the component state. Some instances of this mode are given in Figure 13. In every four-minute vibration video, we extracted 1200 displacement datasets and derived 300 modal sets, and they are then used to estimate the node potentials and edge potentials by GMM and normalized mutual information, respectively.

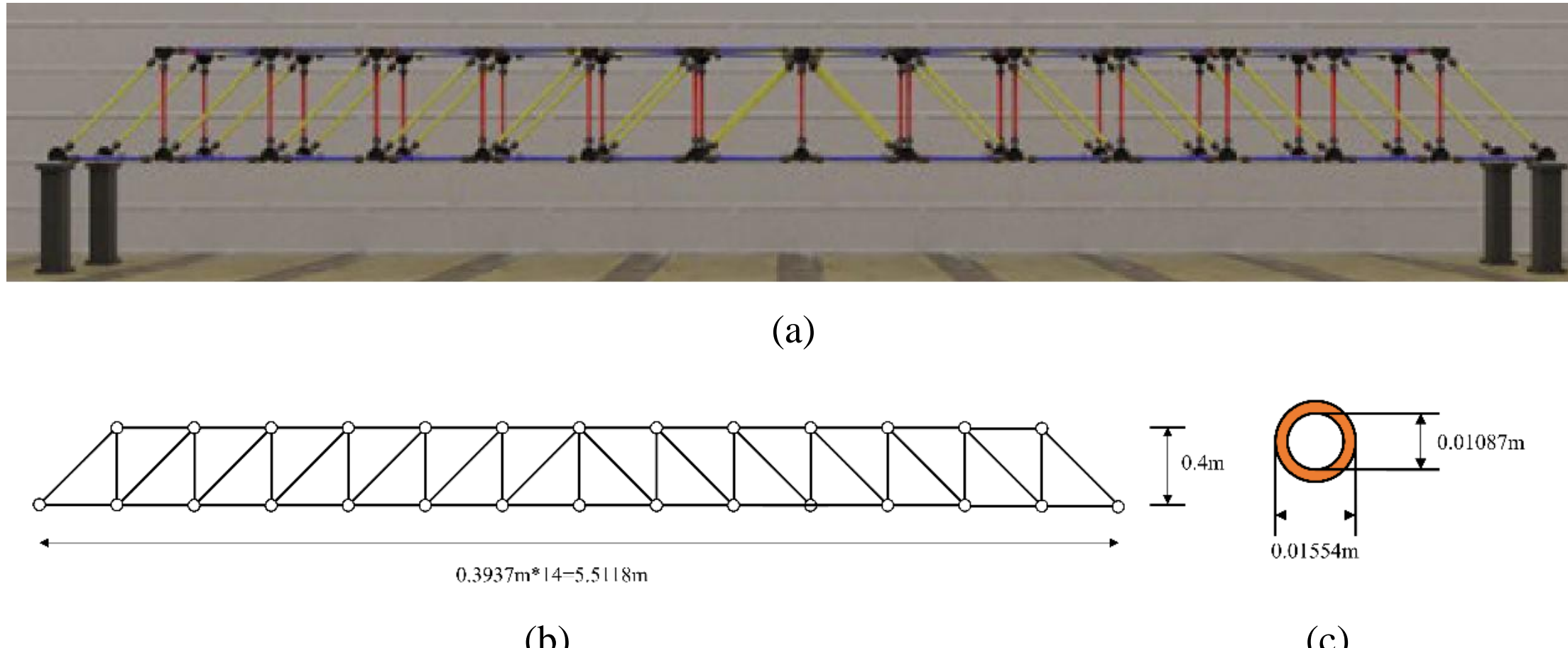

(a)

(b) (c)

Figure 10. An overview of the synthetic truss structure. (a) The main structure. (b) The layout of the structure. (c) The cross section.

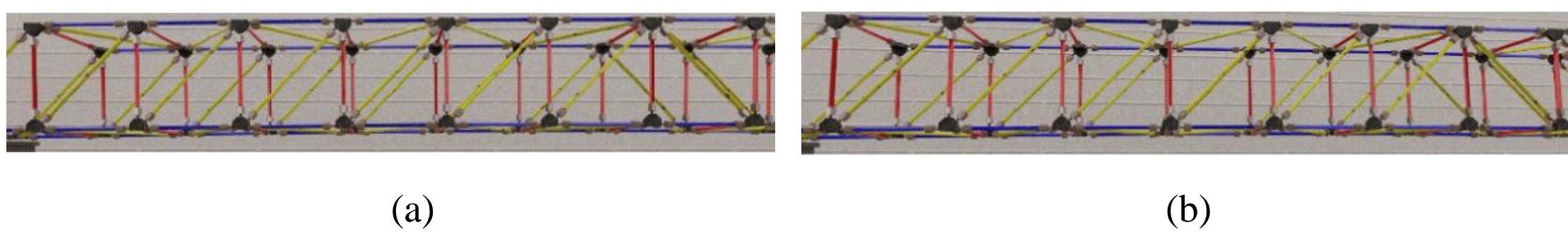

(a) (b)

Figure 11. Two example frames of vibration videos. (a) intact state. (b) damaged state

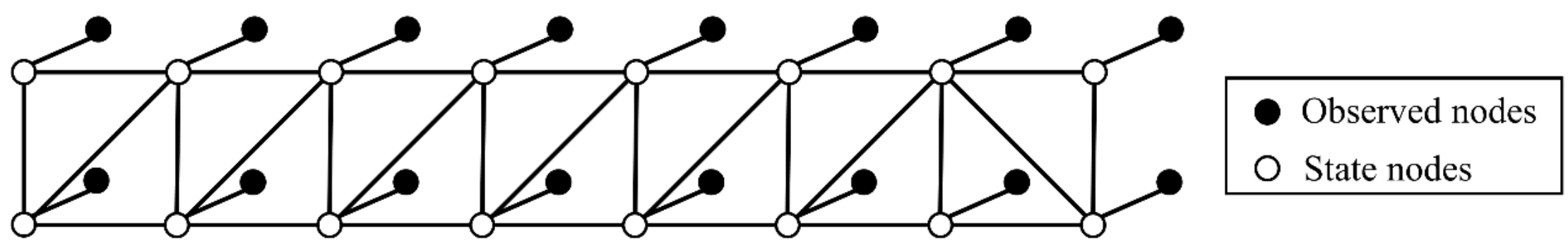

Figure 12. The undirected graphical model constructed for the truss structure based on the proposed structural learning method.

To make accurate inference on the target graphical models, we trained a GNN with strong size generalization performance based on the analysis in the last section. 2000 graphical models with a graph order of 10 were used to train the GNNs, and the average unique node degree of the training graphs is the same as the target graphical model. Six algorithms, including VE, BP, tree-reweighted BP (TRBP) [48], MCMC, variational inference (VI) [21] and GNN are used to infer the component states. VE provides the real probability of damage under the proposed framework, and other methods are all approximate algorithms. Structural components are classified as damaged when their probability exceeds a threshold of 0.5. The specific damage location of each case as well as the inferring results are shown in Figure 14.

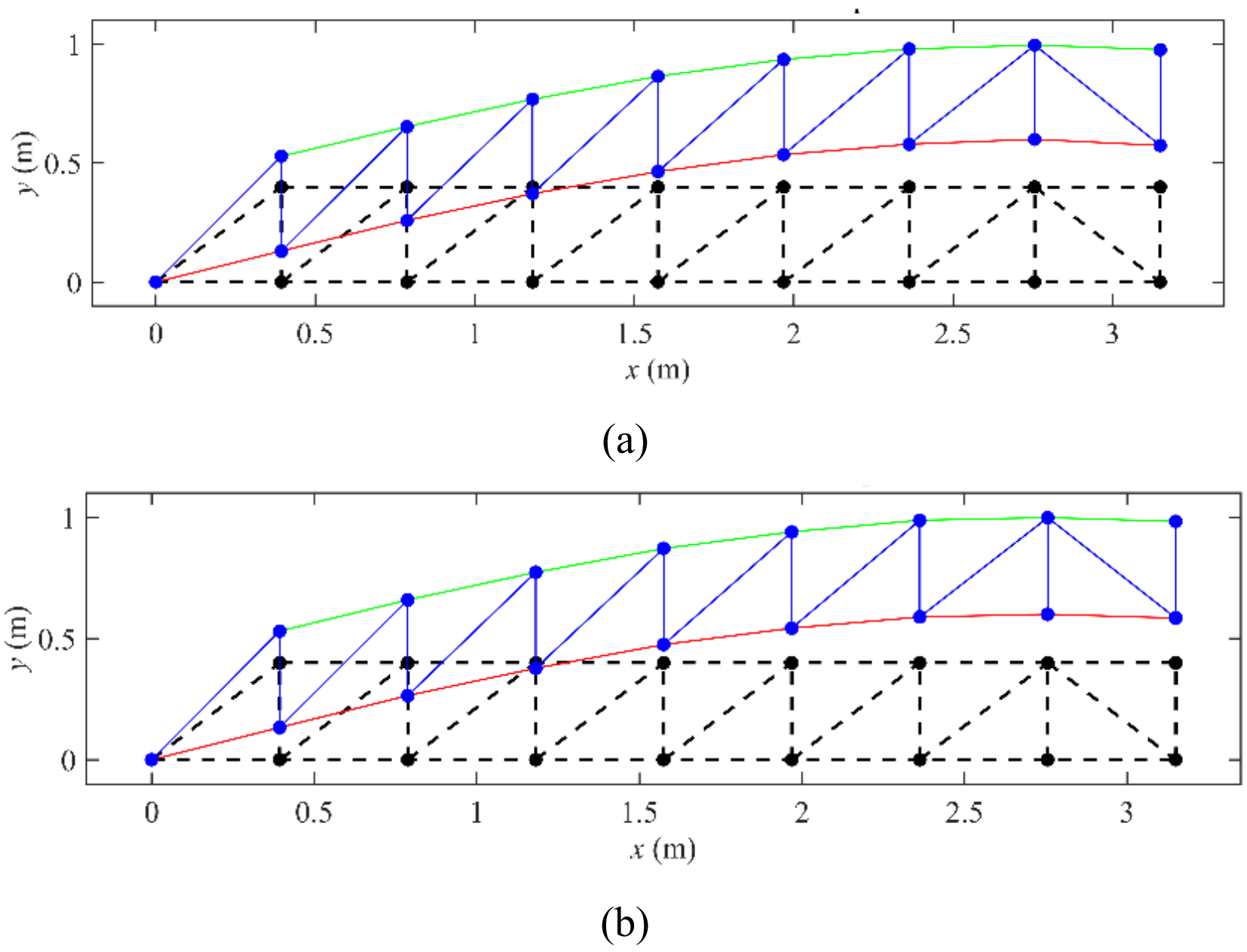

Figure 13. Instances of the fundamental structural mode. Each structural mode is calculated based on measurements from different cases. (a) intact case. (b) damaged case 2.

In order to quantify and compare the performance of each algorithm in inferring the state of components, we have provided four metrics, including the false positive rate (FPR), F1 score [15], accuracy and runtime. Here, we use the state of the truss joints as a basis to indirectly determine the state of these bars. Specifically, we define a node set named “damaged set”, the nodes in this set include the two joints at the ends of the damaged bar. A false positive is made if any node classified as damaged falls outside the “damaged set”, and a false negative is made if the nodes at the ends of the damaged bar are missed in the “damaged set”. The true positives and true negatives are also defined in a similar way. Subsequently, F1 score and accuracy can be computed, and FPR can be calculated by the following equation:

$$\mathrm{FPR} = \frac{|L \cap L_c|}{|L_c|} \tag{30}$$

where $L$ denotes the set of nodes that are classified as damaged; $L_c$ represents the set of all nodes except the two nodes at the end of the damaged bar; $|\cdot|$ is a function for counting the cardinality of a set. The results of four quantitative metrics across different scenarios are shown in Table 2.

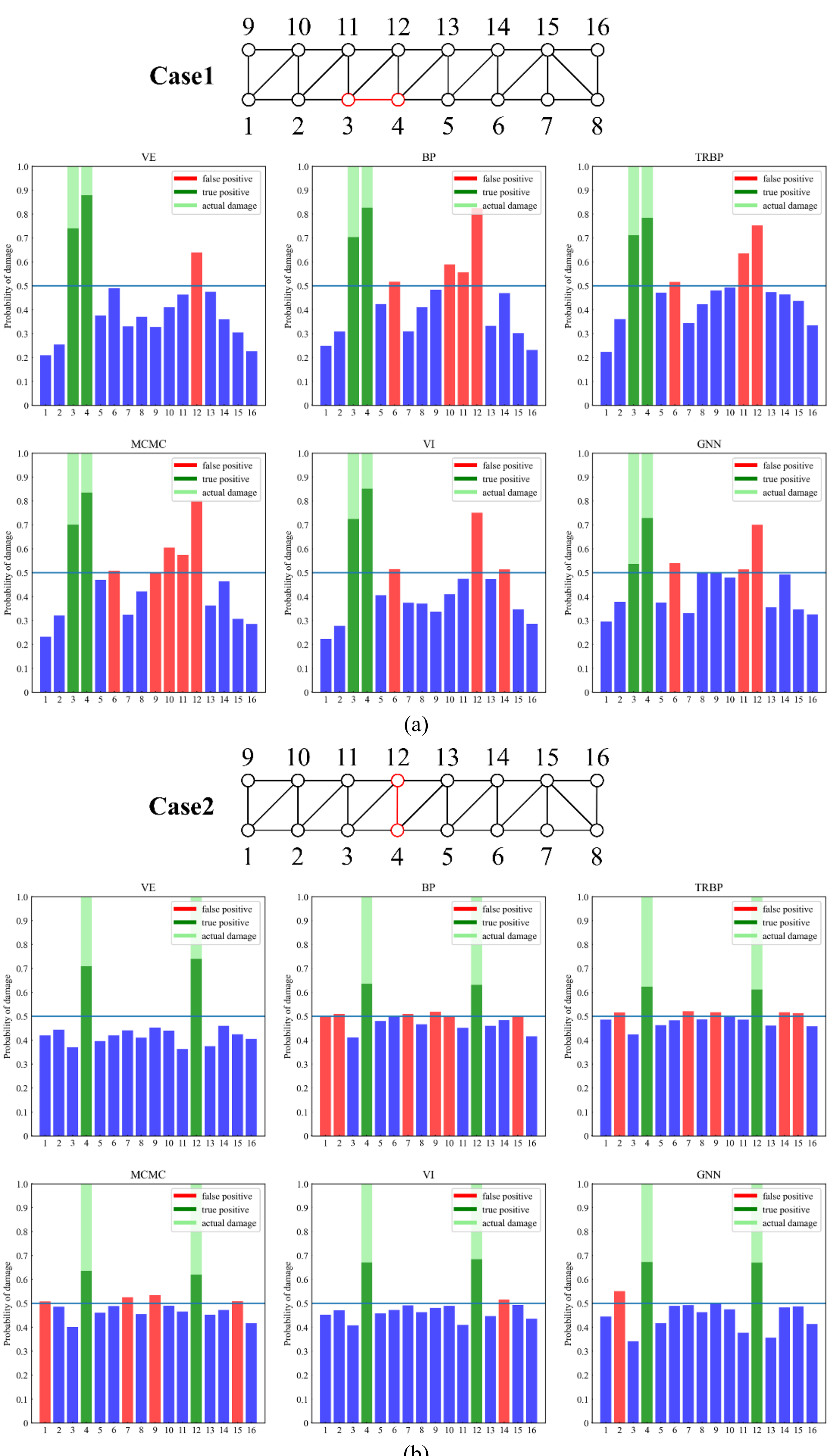

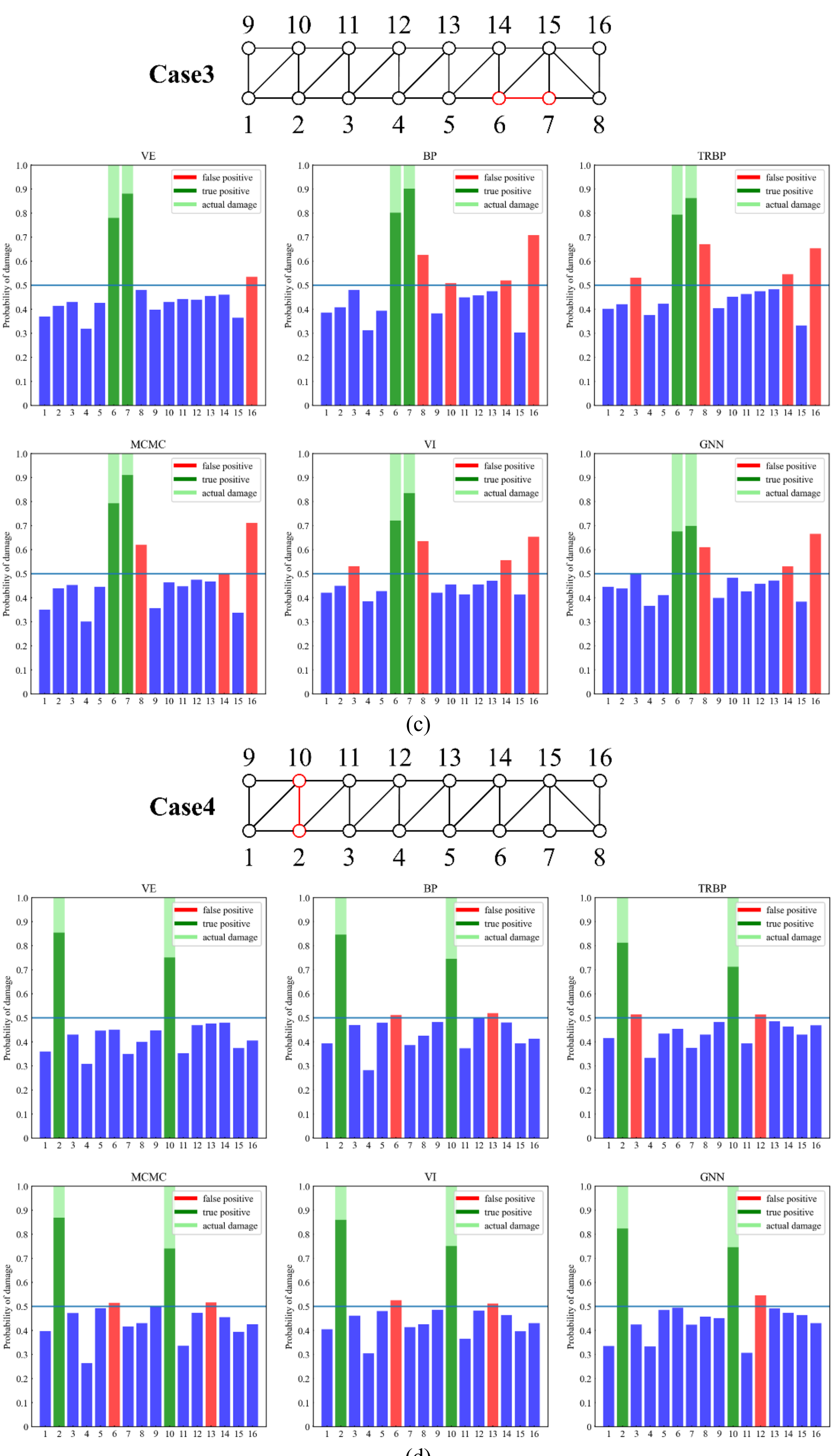

(c)

(d)

Figure 14. The inference results of 4 damage cases. The performance of six algorithms are compared, including VE (exact algorithm, others are approximate algorithms), BP, TRBP, MCMC, VI, and GNN. The vertical axis represents the probability of damage at each node, and the red, green, light green represents false positive, true positive and actual damage, respectively. (a) Damage case 1, damage occurs at the edge connecting 3 and 4. (b) Damage case 2, damage occurs at the edge connecting 4 and 12. (c) Damage case 3, damage occurs at the edge connecting 6 and 7. (d) Damage case 4, damage occurs at the edge connecting 2 and 10.

| | Case 1 | | | | Case 2 | | | |
|---|---|---|---|---|---|---|---|---|
| | FPR | F1 score | accuracy | runtime(s) | FPR | F1 score | accuracy | runtime(s) |
| VE | 0.0714 | 0.8 | 0.9375 | 0.3469 | 0 | 1 | 1 | 0.3466 |
| BP | 0.2857 | 0.5 | 0.75 | 0.0288 | 0.4286 | 0.4 | 0.625 | 0.0292 |
| TRBP | 0.2142 | 0.5714 | 0.8125 | 0.0959 | 0.3571 | 0.4444 | 0.6875 | 0.0986 |
| MCMC | 0.3571 | 0.4444 | 0.6875 | 0.1126 | 0.2857 | 0.5 | 0.75 | 0.1140 |
| VI | 0.2142 | 0.5714 | 0.8125 | 0.0147 | 0.0714 | 0.8 | 0.9375 | 0.0142 |
| GNN | 0.2142 | 0.5714 | 0.8125 | 0.0045 | 0.0714 | 0.8 | 0.9375 | 0.0046 |
| | Case 3 | | | | Case 4 | | | |
| | FPR | F1 score | accuracy | runtime(s) | FPR | F1 score | accuracy | runtime(s) |
| VE | 0.0714 | 0.8 | 0.9375 | 0.3394 | 0 | 1 | 1 | 0.3520 |
| BP | 0.2857 | 0.5 | 0.75 | 0.0285 | 0.1429 | 0.6667 | 0.875 | 0.0292 |
| TRBP | 0.2857 | 0.5 | 0.75 | 0.0928 | 0.1429 | 0.6667 | 0.875 | 0.0984 |
| MCMC | 0.2142 | 0.5714 | 0.8125 | 0.1098 | 0.1429 | 0.6667 | 0.875 | 0.1144 |
| VI | 0.2857 | 0.5 | 0.75 | 0.0143 | 0.1429 | 0.6667 | 0.875 | 0.0149 |
| GNN | 0.2142 | 0.5714 | 0.8125 | 0.0044 | 0.0714 | 0.8 | 0.9375 | 0.0045 |

| Mean value | VE | BP | TRBP | MCMC | VI | GNN |
|---|---|---|---|---|---|---|
| FPR | 0.0357 | 0.2857 | 0.25 | 0.25 | 0.1786 | 0.1429 |
| F1 score | 0.9 | 0.5167 | 0.5456 | 0.5456 | 0.6523 | 0.6857 |
| accuracy | 0.9688 | 0.75 | 0.7813 | 0.7813 | 0.8438 | 0.875 |
| runtime(s) | 0.3462 | 0.0289 | 0.0964 | 0.1127 | 0.0145 | 0.0045 |

Table 2. Four quantitative metrics of inference results in Figure 14.

These results indicate that GNN has lower FPR, higher F1 score and higher accuracy than other approximate algorithms under these cases. Furthermore, GNN spends the least time in inferring the graphical models, which shows its significant computational efficiency. In summary, the GNN-based method demonstrates superior inference accuracy and computational efficiency compared to other approximate algorithms. We have also provided some experiments in Appendix C. Extended experiments to further investigate how the dependencies and special damage scenarios affect the inference results. These experiments can provide a more comprehensive evaluation of the methods proposed in this study

## 6. Conclusion

This work proposes a Bayesian inversion method for discrete structural component states based on PGMs. In contrast to general Bayesian inversion frameworks for continuous parameter variables, this work infers discrete state variables, and leverages PGMs to address the accompanying challenges. By factorizing the joint distribution as a product of potential functions, PGMs can systematically encode the dependencies between structural component states, and the relationships between component state and data. This new paradigm also allows for efficient posterior estimation using a GNN-based algorithm without the need to establish an analytical likelihood function and calculate the complicate integration of the marginal likelihood. Furthermore, to better apply this framework to high-dimensional problems, we systematically investigated the influence of graph properties on GNNs' size generalization capabilities to develop the training strategies that optimize model transferability across varying graph sizes. Our synthetic experiments demonstrated that the average unique node degree and graph order significantly influence the generalization performance of the GNNs. By aligning the properties of training graphs and target graphs, the framework achieves robust inference accuracy on large-scale models with minimal computational overhead. Comparative analyses revealed that GNNs consistently outperform BP and MCMC in both computational efficiency and accuracy. The case studies on truss structures further validated the framework's practical efficacy, and the results indicate that GNNs can achieve better performance than other approximate algorithms in inferring the structural component states.

The proposed method offers a promising pathway for real-world state inversion of structural components, where incomplete datasets and high-dimensional structural models pose persistent challenges. By combining domain knowledge encoded in PGMs with the scalability of GNNs, this framework bridges the gap between probabilistic reasoning and deep learning, enhancing diagnostic reliability and operational feasibility.

### 6.1 Limitations and Future Work

Although this approach achieves good performance within its defined scope, there are still several limitations owing to the simplifications and assumptions. Firstly, the method is designed for the Bayesian inversion of discrete component states. The binary representation is employed, and this simplification is suitable for rapid and preliminary safety screening and significantly enhances computational tractability in high-dimensional problems. However, it cannot precisely quantify performance loss. Moreover, the effectiveness of this binary classification can be challenged under environmental and operational variabilities, which may obscure the true structural state; Secondly, the current framework presents a static model, which may suffer from the dynamic nature of real-world structures; Thirdly, the pairwise Markov network restricts dependencies to directly adjacent components. While the message-passing mechanism in GNNs implicitly propagates information across the graph, this model may not fully capture significant long-range or global couplings present in some complex structures; Furthermore, while the graph property-based training approach can achieve scalability to some extent, training models for exceptionally large infrastructures still demands substantial computational resources to ensure accuracy. These limitations restrict the applicability of the proposed method in certain scenarios, but they also provide directions for future improvements.

Future research will focus on several key directions for extension of the current method: developing a PGM framework capable of handling multi-class state variables to broaden the method's applicability; extending the PGM framework to a spatio-temporal formulation, such as coupling it with a state-space model, thereby enhancing its capability for problems under temporal variations; developing a hierarchical PGM framework that could capture long-range correlations more efficiently.

## 7. Acknowledgment

The study was supported by National Key Research and Development Program of China under Grant No. 2021YFF0501003.

## Appendix A. A simple proof for the equivalence of Bayesian inference and PGM inference

**Theorem 1.** *Let* $p_{M_B}$ *be the posterior distribution derived from a Bayesian inference model* $M_B$*, and* $p_{M_M}$ *be the posterior distribution inferred by a Markov network model* $M_M$*. Then* $p_{M_B}(\boldsymbol{\theta}|\boldsymbol{D}) = p_{M_M}(\boldsymbol{\theta}|\boldsymbol{D})$ *for all* $\boldsymbol{D}$ *such that* $p(\boldsymbol{D}) > 0$ *if and only if the unnormalized product of potential functions defined over clique* $C$ *is proportional to the product of prior distribution and likelihood function, i.e.* $\prod_{c\in C} \psi_c(\boldsymbol{\theta}, \boldsymbol{D}) \propto p(\boldsymbol{\theta})p(\boldsymbol{D}|\boldsymbol{\theta})$

*Proof.*

"⇒." Assume that $\prod_{c\in C} \psi_c(\boldsymbol{\theta}, \boldsymbol{D}) = k \cdot p(\boldsymbol{\theta})p(\boldsymbol{D}|\boldsymbol{\theta})$, then:

$$p_{M_M}(\boldsymbol{\theta},\boldsymbol{D}) = \frac{k}{Z}\cdot p(\boldsymbol{\theta})p(\boldsymbol{D}|\boldsymbol{\theta}) \tag{A-1}$$

and

$$p_{M_M}(\boldsymbol{D}) = \frac{k}{Z}\cdot \sum\nolimits_{\boldsymbol{\theta}} p(\boldsymbol{\theta})p(\boldsymbol{D}|\boldsymbol{\theta}) = \frac{k}{Z} \tag{A-2}$$

So the posterior of the Markov network model is:

$$p_{M_M}(\boldsymbol{\theta}|\boldsymbol{D}) = \frac{p_{M_M}(\boldsymbol{\theta},\boldsymbol{D})}{p_{M_M}(\boldsymbol{D})} = \frac{\frac{k}{Z}\cdot p(\boldsymbol{\theta})p(\boldsymbol{D}|\boldsymbol{\theta})}{\frac{k}{Z}\cdot p(\boldsymbol{D})} = \frac{p(\boldsymbol{\theta})p(\boldsymbol{D}|\boldsymbol{\theta})}{p(\boldsymbol{D})} = p_{M_B} \tag{A-3}$$

"⇐." Assume that $p_{M_B}(\boldsymbol{\theta}|\boldsymbol{D}) = p_{M_M}(\boldsymbol{\theta}|\boldsymbol{D})$

This implies:

$$\frac{p_{M_M}(\boldsymbol{\theta},\boldsymbol{D})}{p_{M_M}(\boldsymbol{D})} = \frac{p(\boldsymbol{\theta})p(\boldsymbol{D}|\boldsymbol{\theta})}{p(\boldsymbol{D})} \tag{A-4}$$

thus,

$$p_{M_M}(\boldsymbol{\theta},\boldsymbol{D}) = \frac{p_{M_M}(\boldsymbol{D})}{p(\boldsymbol{D})}p(\boldsymbol{\theta})p(\boldsymbol{D}|\boldsymbol{\theta}) \tag{A-5}$$

Since $p_{M_M}(\boldsymbol{D})$ and $p(\boldsymbol{D})$ are both functions that only depend on $\boldsymbol{D}$, let $k(\boldsymbol{D}) = \frac{p_{M_M}(\boldsymbol{D})}{p(\boldsymbol{D})}$. Because $p_{M_M}(\boldsymbol{\theta},\boldsymbol{D}) = \frac{1}{Z}\prod_{c\in C}\psi_c(\boldsymbol{\theta},\boldsymbol{D})$, we have:

$$\prod\nolimits_{c\in C}\psi_c(\boldsymbol{\theta},\boldsymbol{D}) = Z\cdot k(\boldsymbol{D})\cdot p(\boldsymbol{\theta})p(\boldsymbol{D}|\boldsymbol{\theta}) \tag{A-6}$$

If $k(\boldsymbol{D})$ is not a constant (i.e. varies with $\boldsymbol{D}$), an additional scaling factor that depends only on $\boldsymbol{D}$ is introduced on the right side. In order for the equation to hold for all $\boldsymbol{\theta}$ and $\boldsymbol{D}$, the function forms on the left and right must match exactly. This requires that $k(\boldsymbol{D})$ must be a constant. Therefore, $\prod_{c\in C}\psi_c(\boldsymbol{\theta},\boldsymbol{D}) \propto p(\boldsymbol{\theta})p(\boldsymbol{D}|\boldsymbol{\theta})$. According to the analysis in **Section 2.2**, the conditions of **Theorem 1** can be easily satisfied, i.e.,

$$\prod_{(i,j)\in\boldsymbol{\varepsilon}}\psi_{ij}(\theta_i,\theta_j)\prod_{i\in\boldsymbol{v}}\phi_i(\theta_i,d_i) = Z\cdot p(\boldsymbol{\theta},\boldsymbol{D}) \propto p(\boldsymbol{\theta})p(\boldsymbol{D}|\boldsymbol{\theta}) \tag{A-7}$$

## Appendix B. General algorithms for PGMs inference

### Appendix B. 1. Variable elimination

Variable elimination is a fundamental exact inference algorithm for computing marginal probabilities in probabilistic graphical models. By systematically eliminating variables through a sequence of sum-product operations, VE efficiently reduces the computational complexity in high-dimensional probability distributions. The algorithm operates as follows: For a given target query $\mathrm{P}(X|E=e)$, where $X$ denotes query variables and $E$ represents observed evidence, VE first identifies an elimination order $\mathcal{Z}$ for non-query, non-evidence variables. For each variable $\mathcal{Z}_i$ in this order, following steps are implemented:

(1) multiplies all factors involving $\mathcal{Z}_i$;

(2) marginalizes $\mathcal{Z}_i$ by summing over its possible states;

(3) replaces these factors with the resulting factor.

This process continues until only the query variables remain, at which point the desired marginal probabilities are obtained. The entire process is given in Algorithm 1.

**Algorithm 1**

Probabilistic graphical model inference through variable elimination

**Input:** Set of factors $\mathcal{F}$ (conditional probability distributions or potentials), Target query variables $X$, Observed evidence variables $E=e$.

$\mathcal{Z}=\{\mathcal{Z}_1,\mathcal{Z}_2,\dots,\mathcal{Z}_n\}$ (Non-query, non-evidence variables)
$\pi=(\pi_1,\pi_2,\dots,\pi_n)$ (Elimination order over $\mathcal{Z}$)
**for** each factor $\phi\in\mathcal{F}$
  **if** $\phi$ contains evidence variables $E$
    reduce $\phi$ by fixing $E=e$
    remove $E$ from the scope of $\phi$
  **end if**
  remove all empty factors from $\mathcal{F}$
**end for**
**for** $i=1\cdots n$ **do**
  $\mathcal{Z}=\pi_1$ (Select the variable to be eliminated)
  $\mathcal{F}_z\leftarrow\{\phi\in\mathcal{F}|\mathcal{Z}\in\mathrm{Scope}(\phi)\}$ (Collect all factors related to current $\mathcal{Z}$)
  $\psi\leftarrow\prod_{\phi\in\mathcal{F}_z}\phi$ (Factor product)
  $\tau\leftarrow\sum_{\mathcal{Z}}\psi$ (Marginalization)
  $\mathcal{F}\leftarrow(\mathcal{F}\backslash\mathcal{F}_{\mathcal{Z}})\cup\{\tau\}$
**end for**
$\beta\leftarrow\prod_{\phi\in\mathcal{F}_{\mathcal{Z}}}\phi$ (Product all factors involving the evidence)
$\beta^*\leftarrow\frac{\beta}{\sum_X\beta}$ (normalization)

**Return** $\beta^*$ (marginal probabilities given the evidence $\mathrm{P}(X|E=e)$)

---

**Appendix B. 2. Belief propagation algorithm**

Belief propagation (BP), also known as sum-product algorithm, provides an efficient framework for marginal probability computation in probabilistic graphical models. This algorithm calculates the messages transmitted between all adjacent nodes, which is expressed as:

$$m_{ij}^{t}(x_j) = \alpha \sum_{x_i} e^{-b_i x_i - \omega_{ij}(x_i x_j - 1)} \prod_{k \in N(i)/j} m_{ki}^{t-1}(x_i) \tag{B-1}$$

Where $m_{ij}(x_j)$ denotes the messages from node $i$ to node $j$; $\alpha$ is a normalization constant ensuring $\sum_{x_i} m_{ij}(x_j) = 1$; and $N(i)$ denotes all the neighboring nodes of node $i$; $\omega_{ij}$ and $b_i$ represents edge potential and node potential, respectively. After message iterates to step $T$, node beliefs are computed as:

$$\mathcal{B}_i(x_i) = \gamma e^{-b_i x_i} \prod_{j \in N(i)} m_{ji}^{T}(x_i) \tag{B-2}$$

The specific process is shown in Algorithm 2.

---

**Algorithm 2**

Probabilistic graphical model inference through belief propagation

---

**Input:** Graph $\mathcal{G}(\nu, \varepsilon)$, edge potential $\omega_{ij}$, node potential $b_i$, tolerance $\epsilon$

$m_{ij}^{0}(x_j) \leftarrow \frac{1}{|\mathcal{X}|}$ (Initialize messages with uniform distribution)

$m_{ji}^{0}(x_i) \leftarrow \frac{1}{|\mathcal{X}|}$

**while** $\max|m^{t+1} - m^{t}| > \epsilon$ **do**

**for** each variable $i \in \nu$ **do**

update $m_{ij}^{t}(x_j)$ by Eq. (B-1)

**end for**

**end while**

Compute marginal distributions $\mathcal{B}_i(x_i)$ by Eq. (B-2)

**Return** $\{\mathcal{B}_i(x_i)\}$

---

**Appendix B. 3. Monte Carlo methods**

Monte Carlo methods provide a computationally feasible framework for approximate inference in probabilistic graphical models where exact computation of marginal probabilities becomes intractable. This kind of method samples from a closely-related distribution and estimates the marginal probabilities by sample-based averages. Here we give a brief introduction of the Gibbs sampling.

The Gibbs sampling approach belongs to the class of Markov Chain Monte Carlo (MCMC) methods. This algorithm iteratively constructs a Markov chain that converges to the target stationary distribution $p(x)$, thereby enabling effective sample generation through chain transitions. Specifically, the Gibbs sampler samples each variable conditioned on its Markov blanket. For a variable $x_i$, let $\mathrm{Mb}(x_i)^{t-1}$ denote its Markov blanket (neighbors of node $i$ in the graph). The conditional distribution satisfies:

$$p(x_i'|x_{-i}) \propto \prod_{c:x_i\in c} \psi_c(x_c) \tag{B-3}$$

Let $x^{(t)}$ denote the state at iteration $t$. The transition kernel for updating is:

$$T\left(x^{(t+1)}\middle|x^{(t)}\right) = \prod_{i=1}^{n} P\left(x_i^{(t+1)}\middle|\mathrm{Mb}(x_i)^t\right) \tag{B-4}$$

The balance condition $\pi(x)T(x'|x) = \pi(x')T(x|x')$ ensures convergence to the stationary distribution. Afterwards, a new value is sampled from this updated distribution and it will replace the previous value. Specific process of Gibbs sampling is shown in Algorithm 3.

---

**Algorithm 3**

Probabilistic graphical model inference through MCMC (Gibbs sampling)

---

**Input:** Graph $\mathcal{G}(\nu, \varepsilon)$, clique potentials $\psi_j(x_j)$ (defined in section 2.1), number of iterations T

$x_i^{(0)} \leftarrow 0$ (Initialize the samples arbitrarily)

**for** $t = 1 \cdots T$ **do**

  **for** each variable $x_i$ **do**

    compute conditional distribution by Eq. (B-4)

    sample $x_i^{(t)} \sim P(x_i|\mathrm{Mb}(x_i)^{t-1})$

  **end for**

**end for**

**Return** $x_i^{(t0)}, \dots, x_i^{(t)}$

---

## Appendix C. Extended experiments

This part presents extended experiments to evaluate the robustness of the proposed Bayesian inversion framework under two conditions. The first is that the physical structure and the adopted graphical model are not matched, and the second is that the dependencies between components are not considered.

### Appendix C.1. Robustness Tests for Model Mismatch

This experiment tests the method's sensitivity when damage occurs in structural components that are not directly represented by the edges or nodes of the learned probabilistic graphical model. This simulates practical scenarios where the model simplification may not capture all potential damage locations. Two simulation cases are considered, and the corresponding graphical models as well as the inference results are shown in Figure 15. The GNNs used to make inference are trained on graphs which have similar properties with the established graphical model. Three quantitative metrics are also computed in Table 3.

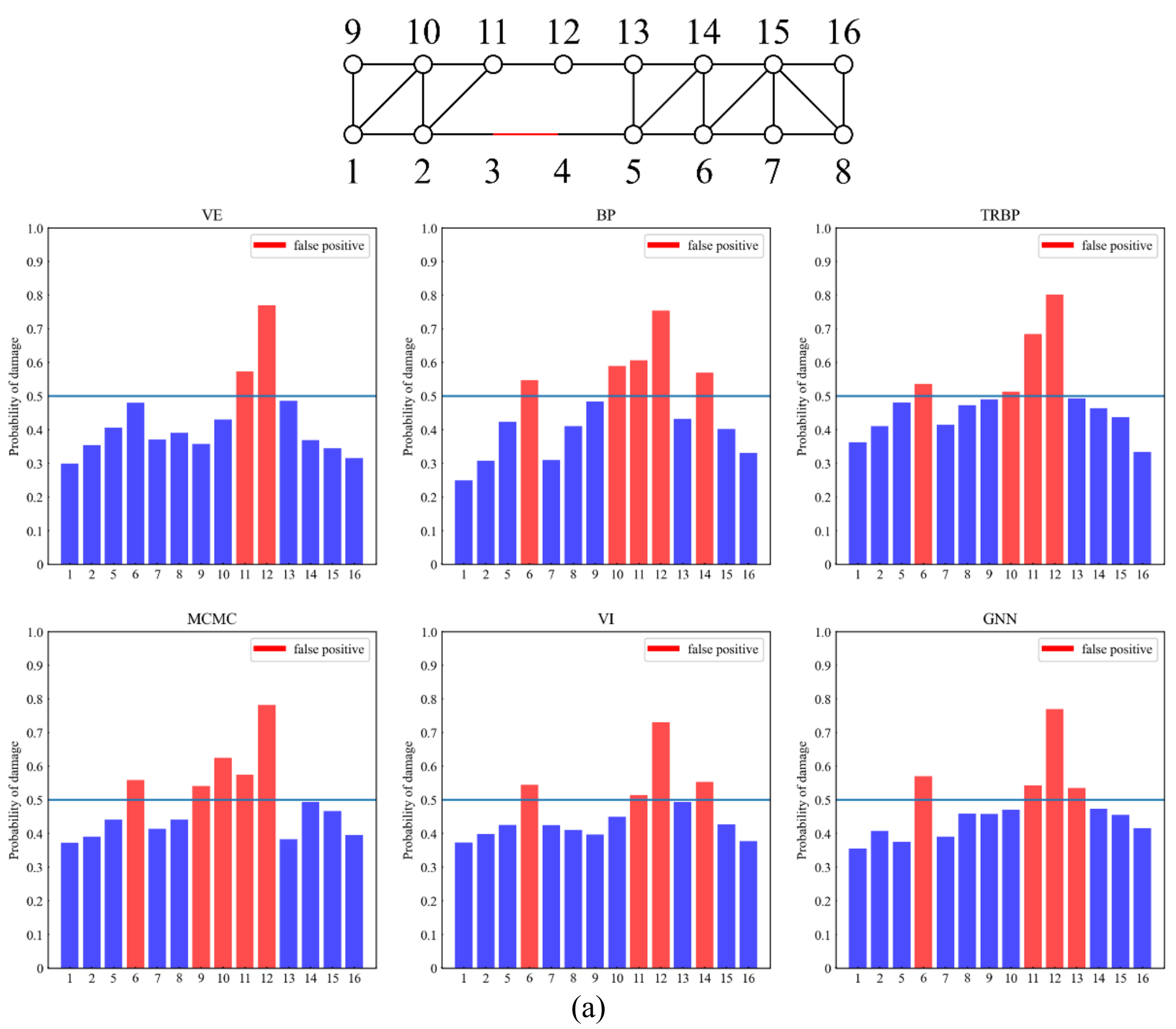

(a)

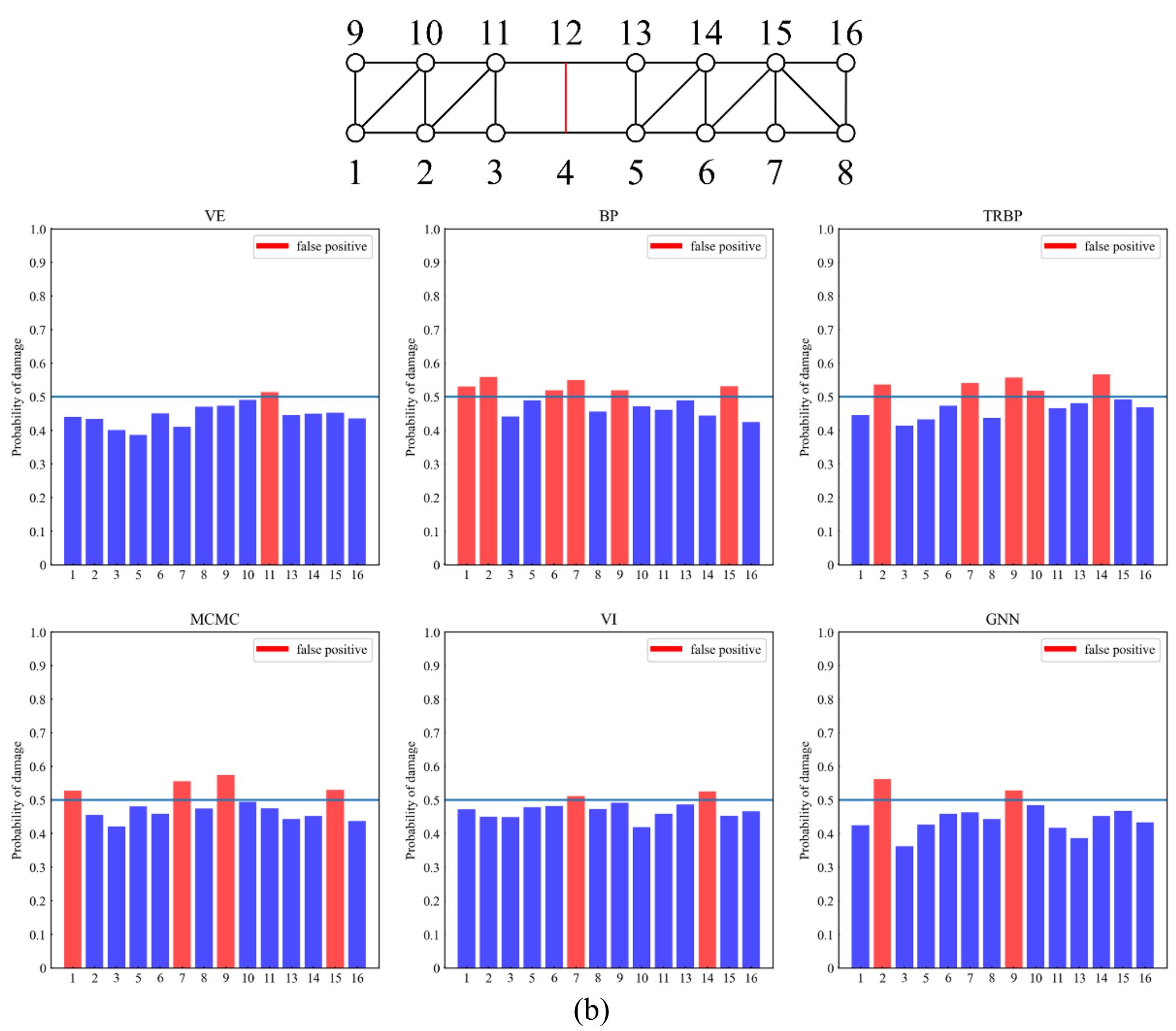

(b)

Figure 15. The inference results of two experiments where damage occurs in structural components that are not represented in the graphical model.

| | Figure 15(a) | | | Figure 15(b) | | |
|---|---|---|---|---|---|---|
| | FPR | F1 score | accuracy | FPR | F1 score | accuracy |
| VE | 0.1429 | 0.6667 | 0.875 | 0.0714 | 0.8 | 0.9375 |
| BP | 0.3571 | 0.4444 | 0.6875 | 0.4286 | 0.4 | 0.625 |
| TRBP | 0.2857 | 0.5 | 0.75 | 0.3571 | 0.4444 | 0.6875 |
| MCMC | 0.3571 | 0.4444 | 0.6875 | 0.2857 | 0.5 | 0.75 |
| VI | 0.2857 | 0.5 | 0.75 | 0.1429 | 0.6667 | 0.875 |
| GNN | 0.2857 | 0.5 | 0.75 | 0.1429 | 0.6667 | 0.875 |

Table 3. The quantitative metrics of the results in Figure 15.

The results in Figure 15 and Table 3 show that when damage occurs in structural components that are not represented in the graphical model, there would be more false positives, which means that the proposed method is somewhat sensitive to such scenarios. Consequently, the complete structure should be modeled as much as possible when establishing the probabilistic graphical model.

### Appendix C.2. Dependency Analysis

The second experiment is to examine the contribution of modeling spatial dependencies between component states to the overall inference accuracy. This is achieved by setting all edge potential parameters to zero, which removes the terms that couple the states of adjacent components. The inference results under the same four damage cases as in Section 5 are shown in Figure 16, and the quantitative metrics are computed in Table 4.

The results show that in Case 2 and 3, the actual damages are not correctly detected (false negatives) when data dependencies are ignored. This might indicate that ignoring dependencies leads to detection failures. In addition, compared with the results in Figure 14, there are more false positives when the dependencies are not take into account. Therefore, the explicit modeling of spatial dependencies is shown to be a key factor improving overall inference accuracy.

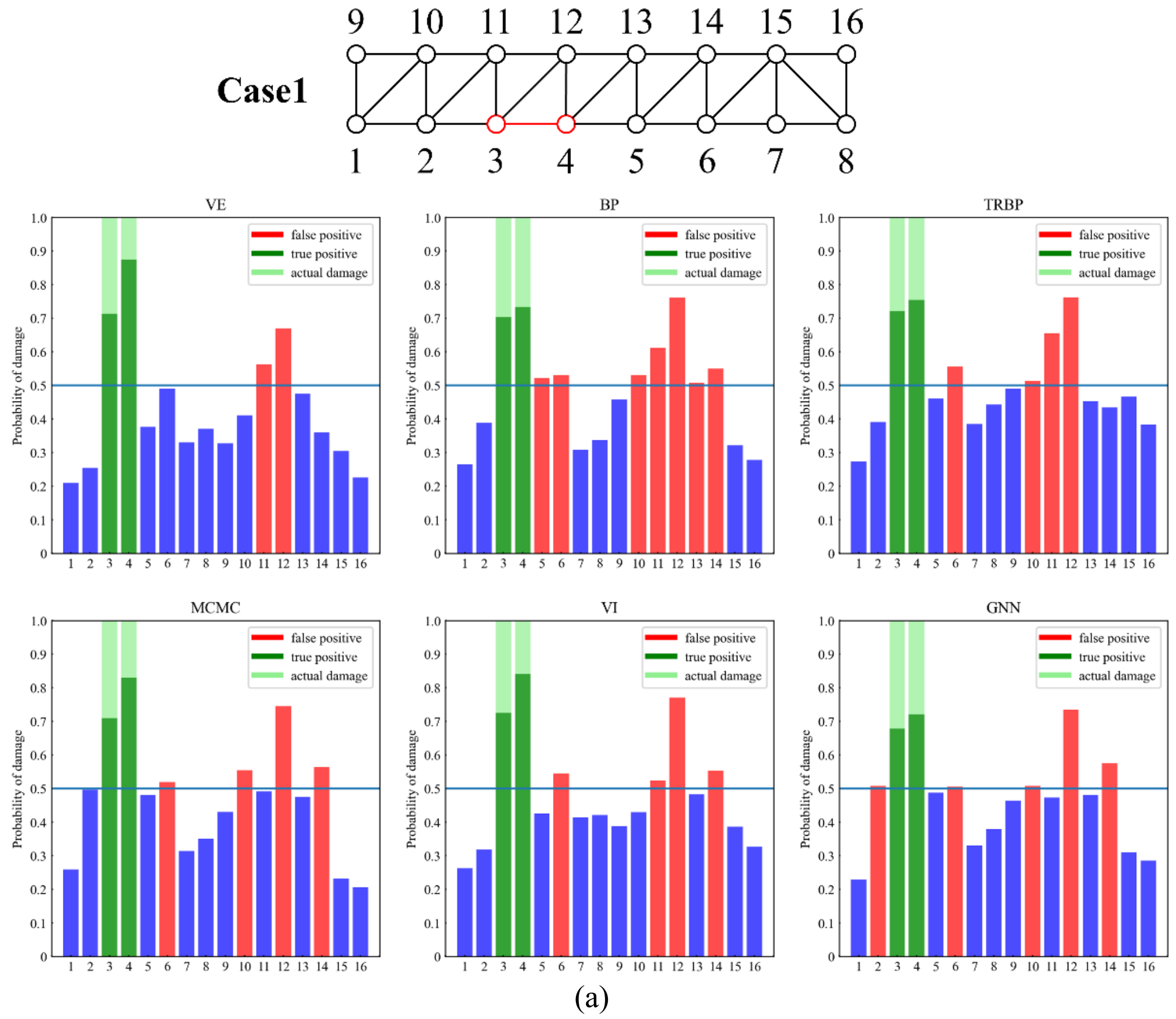

(a)

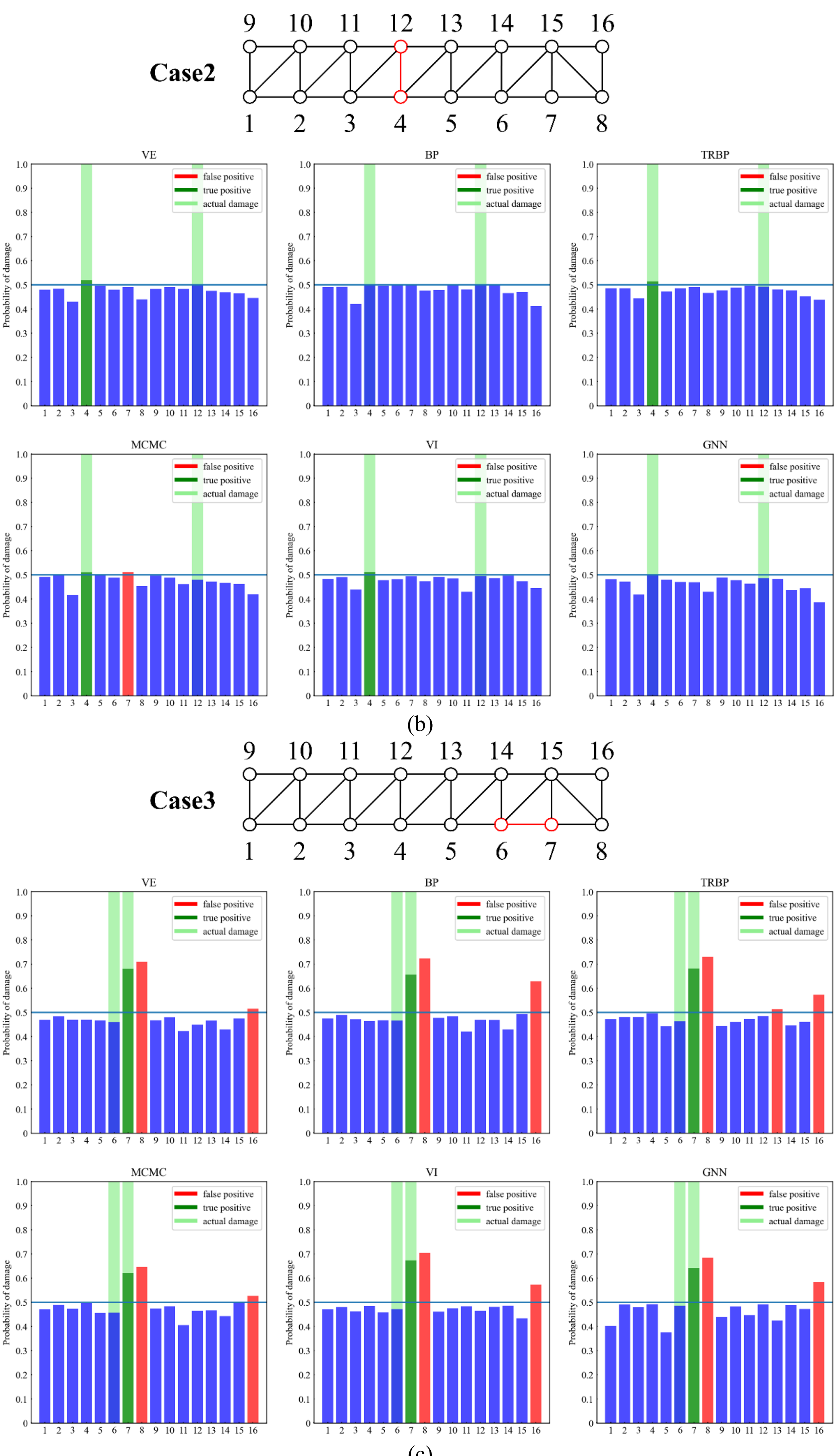

(b)

(c)

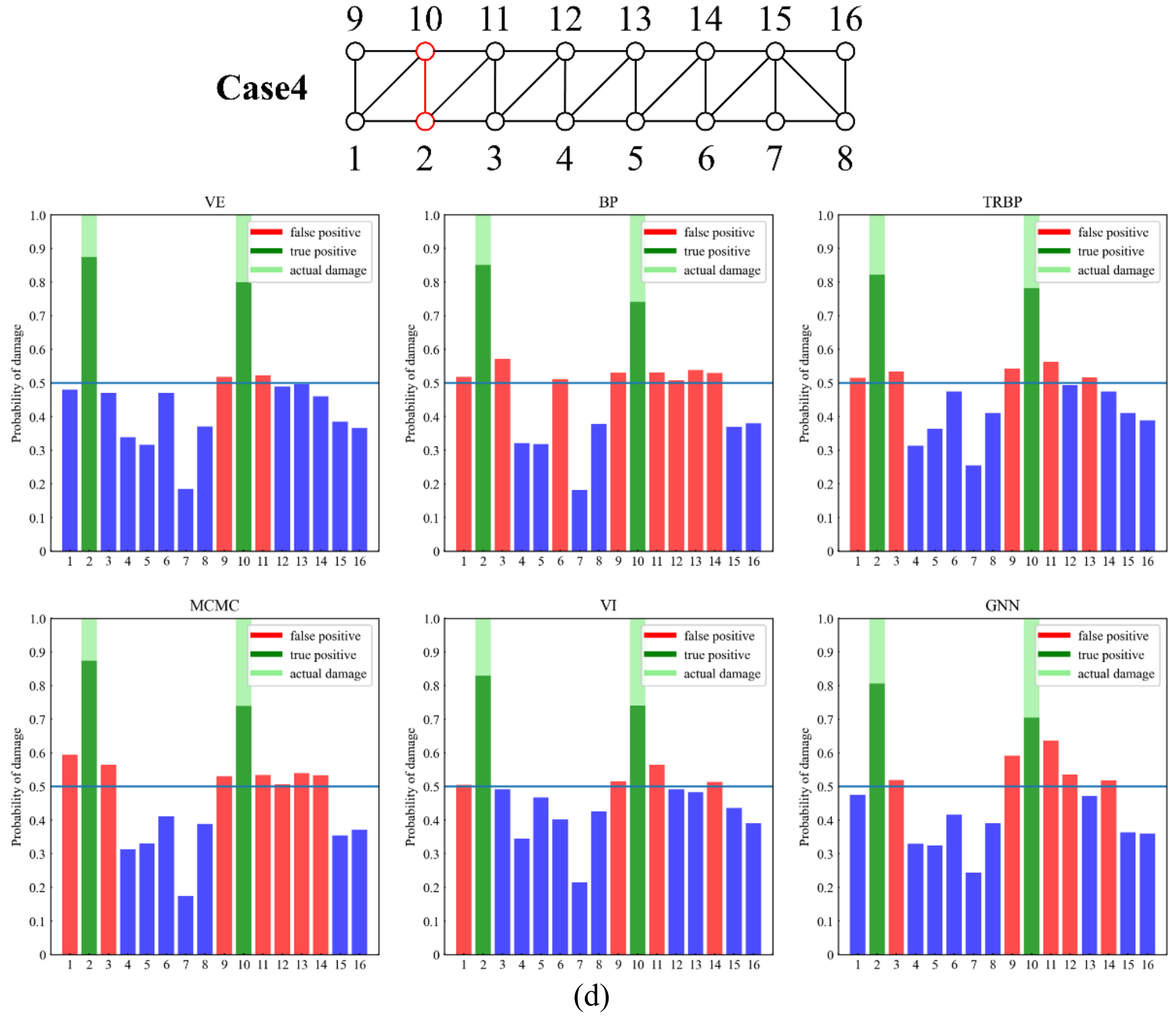

(d)

Figure 16. The inference results of four damage cases. The dependencies between component states are not considered. It can be found that there are false negatives in Case 2 and 3, this might indicate that ignoring dependencies leads to detection failures.

| | Case 1 | | | Case 2 (false negative) | | |
|---|---|---|---|---|---|---|
| | FPR | F1 score | accuracy | FPR | F1 score | accuracy |
| VE | 0.1429 | 0.6667 | 0.875 | 0 | 0.6667 | 0.9375 |
| BP | 0.5 | 0.3636 | 0.5625 | 0 | 0 | 0.875 |
| TRBP | 0.2857 | 0.5 | 0.75 | 0 | 0.6667 | 0.9375 |
| MCMC | 0.2857 | 0.5 | 0.75 | 0.0714 | 0.5 | 0.875 |
| VI | 0.2857 | 0.5 | 0.75 | 0 | 0.6667 | 0.9375 |
| GNN | 0.3571 | 0.4444 | 0.6875 | 0 | 0 | 0.875 |
| | Case 3 (false negative) | | | Case 4 | | |
| | FPR | F1 score | accuracy | FPR | F1 score | accuracy |
| VE | 0.1429 | 0.4 | 0.8125 | 0.1429 | 0.6667 | 0.875 |
| BP | 0.1429 | 0.4 | 0.8125 | 0.5714 | 0.3333 | 0.5 |
| TRBP | 0.2143 | 0.4 | 0.8125 | 0.3571 | 0.4444 | 0.6875 |
| MCMC | 0.1429 | 0.3333 | 0.75 | 0.5 | 0.3636 | 0.5625 |
| VI | 0.1429 | 0.5 | 0.875 | 0.2857 | 0.5 | 0.75 |
| GNN | 0.1429 | 0.4 | 0.8125 | 0.3571 | 0.4444 | 0.6875 |

Table 4. The quantitative metrics of the results in Figure 16.